\documentclass[preprint]{elsarticle}
\biboptions{longnamesfirst}
\journal{Expert Systems with Applications}

\usepackage[utf8]{inputenc} % allow utf-8 input
\usepackage[T1]{fontenc}    % use 8-bit T1 fonts
\usepackage{hyperref}       % hyperlinks
\usepackage{url}            % simple URL typesetting
\usepackage{booktabs}       % professional-quality tables
\usepackage{nicefrac}       % compact symbols for 1/2, etc.
\usepackage{microtype}      % microtypography
\usepackage{lipsum}

\usepackage{graphicx}
\usepackage{subcaption}
\usepackage{caption}

\usepackage{amsfonts}
\usepackage{amsmath}
\usepackage{amssymb}
\usepackage{amsthm}

\usepackage{tikz}
\usetikzlibrary{shapes.geometric, arrows, calc, fit}
\begin{document}
\begin{frontmatter}

% Title
\title{Graph Signal Recovery Using Restricted Boltzmann Machines}

% Author names and emails
\author[usc]{Ankith Mohan}
\ead{ankithmo@usc.edu}

\author[usc]{Aiichiro Nakano}
\ead{anakano@usc.edu}

\author[isi]{Emilio Ferrara\corref{cor}}
\cortext[cor]{Corresponding author}
\ead{ferrarae@isi.edu}

% Addresses
\address[usc]{
    Department of Computer Science,
    University of Southern California,
    Los Angeles\\
    CA, USA 90089
}
\address[isi]{
    Information Sciences Institute,
    University of Southern California,
    Marina del Rey\\
    CA, USA 90292
}

%%%%%%%%%%%%%%%%%%%%%%%%%%%%%%%%%%%%%%%%%%%%%%%%%%%%%%%%%%%%%%
% Abstract
%%%%%%%%%%%%%%%%%%%%%%%%%%%%%%%%%%%%%%%%%%%%%%%%%%%%%%%%%%%%%%
\begin{abstract}
We propose a model-agnostic pipeline to recover graph signals from an expert system by exploiting the {\it content addressable memory} property of restricted Boltzmann machine and the representational ability of a neural network.
The proposed pipeline requires the deep neural network that is trained on a downward machine learning task with {\it clean} data, data which is free from any form of corruption or incompletion.
We show that denoising the representations learned by the deep neural networks is usually more effective than denoising the data itself.
Although this pipeline can deal with noise in any dataset, it is particularly effective for graph-structured datasets.
\footnote{Code is available at \url{ https://github.com/ankithmo/denoiseRBM}.}
\end{abstract}
%%%%%%%%%%%%%%%%%%%%%%%%%%%%%%%%%%%%%%%%%%%%%%%%%%%%%%%%%%%%%%
% Keywords
%%%%%%%%%%%%%%%%%%%%%%%%%%%%%%%%%%%%%%%%%%%%%%%%%%%%%%%%%%%%%%

%CRediT authorship contribution statement:
%\textbf{Ankith Mohan:} 
%Conceptualization, 
%Methodology, 
%Software,
%Validation,
%Formal analysis,
%Investigation,
%Resources,
%Data curation,
%Writing - Original Draft,
%Writing - Review \& Editing,
%Visualization.
%\textbf{Aiichiro Nakano:}
%Resources,
%Writing - Review \& Editing,
%Visualization,
%Supervision,
%Project administration.
%\textbf{Emilio Ferrara:}
%Conceptualization,
%Methodology,
%Resources,
%Writing - Review \& Editing,
%Visualization,
%Supervision,
%Project administration.

%Declaration of Competing Interest:
%The authors declare that they have no known competing financial interests or personal relationships that could have appeared to influence the work reported in this paper.

%\maketitle
\end{frontmatter}
%%%%%%%%%%%%%%%%%%%%%%%%%%%%%%%%%%%%%%%%%%%%%%%%%%%%%%%%%%%%%%
% Introduction [Incomplete. Requires a lot of polishing]
%%%%%%%%%%%%%%%%%%%%%%%%%%%%%%%%%%%%%%%%%%%%%%%%%%%%%%%%%%%%%%
%\begin{document}
\section{Introduction}\label{sec:intro}
% Graphs and graph-structured datasets
Graphs are a ubiquitous data structure that are employed extensively in almost every field of study.
This is due to their ability to efficiently store complex information in a simple manner that is amenable to mathematical analysis.
Social, information, biological, ecological and recommendation networks are just a few examples of the fields that can be readily modeled as graphs, which capture interactions between individuals.
The properties of the nodes, the existence and properties of the edges are efficiently organized into a {\it graph-structured} dataset.

% Machine learning using such graph signals
However, graphs are not only useful as structured knowledge repositories, they also play a pivotal role in modern
machine learning (ML).
Many ML applications seek to make predictions or identify patterns using these graph signals. 
For instance, predicting the role of a person in a collaboration network, recommending new content to a user in a social network, or predicting new applications of molecules, all of which can be represented as graphs.

% Graph neural networks and graph signals
{\it Neural networks} (NN) have demonstrated strong power in learning abstract, yet effective features from a huge amount of data~\cite{nielsen2015neural}. 
As the extension of NNs to the graph domain, {\it graph neural networks} (GNNs) have also received a lot of attention and achieved significant success in solving ML problems on graph-structured data~\cite{thekumparampil2018attention,chiang2019cluster,velivckovic2017graph,kipf2016semi,xu2018powerful,hamilton2017inductive}. These GNNs attempt to learn low dimensional representations which capture the geometric dependencies between nodes in the graph. 
Alternatively, we can consider the graph-structured data and the learned representations as signals on the underlying graph.

% Signal recovery problem
{\it Signal recovery} recovers one or multiple smooth
signals from corrupted, or incomplete measurements.
These problems are usually concerned with image denoising and signal inpainting.
% Image denoising
{\it Image denoising} aims to reconstruct a high quality image from its degraded observation.
\cite{gu2019brief} provides a review of prior modeling
approaches, conventional sparse representation based denoising algorithms, low-rank based denoising algorithms and recent deep neural networks based approaches.
\cite{veerakumar2019empirical} proposes a system to identify and correct images affected by impulse noise.
The system has two steps:
(1) noise pixel identification using empirical mode decomposition, and
(2) restoration of the noisy pixels using adaptive bilateral filtering.
The system is also capable of preserving the edges and performs very well for high Gaussian noisy environments.

% Signal inpainting
{\it Signal inpainting} reconstructs lost or deteriorated parts of signals, including images and videos. \cite{elharrouss2019image} summarizes current image inpainting techniques into sequential-based, CNN-based and GAN-based methods.
\cite{abraham2012survey} surveys patch-based and object-based video inpainting techniques.

% Graph signal recovery problem
When we are dealing with signals on an underlying graph, we are posed with the {\it graph signal recovery problem}.
This is particularly problematic because the distortions are not only possible in the properties of the nodes but also in the existence of edges as well as their properties.

% denoiseRBM
We propose a denoising pipeline for alleviating the graph signal recovery problem, by exploiting the property of restricted Boltzmann machines (RBMs) that can act as a content-addressable memory, and the representational ability of NNs. 
Figure~\ref{fig:denoise} illustrates the proposed pipeline. This denoising pipeline can work with any deep neural network (DNN) trained on downward machine learning (ML) task with {\it clean} training data, data that is free from corruption and incompletion.

We briefly describe the pipeline in the following steps, while a detailed explanation is provided in section~\ref{sec:denoise}.
\begin{enumerate}
    \item We assume that we are provided with a trained DNN, $\phi$.
    \item Given $\phi$, we first choose the hidden layer $i$ whose representations $z_{i}$ we find most informative for the denoising task.
    \item We train an RBM on $z_{i}$, denoted by {\it RBM-$z_{i}$}. The denoising pipeline with this RBM is denoted as $\psi_{i}$.
    \item We pass the data with $n\%$ noise through $\psi_{i}$ to obtain some representation at layer $i$, $z_{i}^{[n]}$ which is a noisy estimate of the true representations that would have been generated with noise-free test data. 
    \item $z_{i}^{[n]}$ is fed to {\it RBM-$z_{i}$} to obtain a denoised representation $\tilde{z_{i}}^{[n]}$, which is then fed to the layer immediately succeeding layer $i$.
    \item We continue processing $\tilde{z_{i}}^{[n]}$ through $\phi$ to obtain the output.
\end{enumerate}

For instance, suppose that the second hidden layer has the most informative representation for our problem. 
Then, we would extract $z_{2}$ from $\phi$ and train {\it RBM-$z_{2}$} on these embeddings. 
This would give us $\psi_{2}$. 
Then we would do the following:
\begin{enumerate}
    \item Pass $n\%$ noisy data into $\psi_{2}$.
    \item Obtain the noisy estimates $z_{2}^{[n]}$.
    \item Feed these estimates to {\it RBM-$z_{2}$}.
    \item Recover reconstructions $\tilde{z_{2}}^{[n]}$.
    \item Push these reconstructions back into the pipeline through $BN_{2}$ (refer figure~\ref{fig:denoise}).
    \item Obtain output.
\end{enumerate}

\paragraph{Outline of the paper} Section~\ref{sec:bg} reviews DNNs and RBMs, which lays the foundation for this paper. 
Section~\ref{sec:motive} formulates the problem and describes the rationale behind our proposed solution. 
Section~\ref{sec:denoise} details our approach. 
We describe our experiments in section~\ref{sec:expts} followed with its results in section~\ref{sec:results}. 
Section~\ref{sec:analysis} focuses on the representative behaviour for our pipelines which are discussed in section~\ref{sec:discuss}.
Section~\ref{sec:related} summarizes the related work while section~\ref{sec:conclude} concludes the paper and provides pointers to future direction.
%%%%%%%%%%%%%%%%%%%%%%%%%%%%%%%%%%%%%%%%%%%%%%%%%%%%%%%%%%%%%%
% Background [Might be a lot]
%%%%%%%%%%%%%%%%%%%%%%%%%%%%%%%%%%%%%%%%%%%%%%%%%%%%%%%%%%%%%%
\section{Background}\label{sec:bg}
In this section, we will introduce some key concepts that form the foundation for the remainder of the paper.

\subsection{Deep neural networks from the perspective of information theory}\label{sec:DNN_MI}
% What do we have?
Let $X$ denote the input variable and $Y$ denote the output variable. 
$(x,y) \sim P(X,Y)$ where $P(X,Y)$ is the unknown true distribution of ($X,Y$).
We use $\mathcal{X} \sim P(X)$ and $\mathcal{Y} \sim P(Y)$ to denote the set of data points and their corresponding labels drawn from their respective distributions. 
Thus, the dataset $\mathcal{D} = \{(x_{i},y_{i})\}_{i=1}^{n} \sim P(X,Y)$ are independent and identically distributed (iid) samples from the joint distribution.

% Supervised learning
Given $\mathcal{D}$, our task is to learn the mapping from $\mathcal{X}$ to $\mathcal{Y}$ so as to maximize the prediction accuracy.
% What is the issue?
This is possible when samples belonging to different classes in $\mathcal{X}$ are linearly separable.
Linear separability requires data points to be conditionally independent given their output classification.
However, we cannot just assume conditional independence for $\mathcal{X}$.

% How do we solve this issue?
$X$ is usually a high dimensional variable which is a low level representation of the data.
When compared with $X$, $Y$ has significantly lower dimensionality of the predicted categories.
Thus, most of the entropy of $X$ is not informative about $Y$.
More importantly, the relevant features in $X$ are highly distributed and difficult to extract.
Deep neural networks (DNNs) are capable of extracting this distributed information by sequentially transforming the representation in its layered structure.

% How do DNNs work?
Given the joint distribution $P(X,Y)$, the relevant information is defined as the {\it mutual information} $I(X;Y)$ where statistical dependence is assumed between $X$ and $Y$.
$Y$ implicitly determines both the relevant and the irrelevant features in $X$.
Optimal representation of $X$ would capture relevant features, and compress $X$ by dismissing irrelevant parts which do not contribute to prediction of $Y$.
The layered structure of DNN result in successive refining of the relevant information in $X$ such that we arrive at this desired optimal representation~\cite{tishby2015deep}.

% How to quantify an DNN?
The advantage of this layered structure is that this results in a Markovian structure and the data processing inequalities ensure that information lost in one layer cannot be recovered in the subsequent layers. 
If $T_{l}, \forall\ l \in [L]$ denotes the representations learned by the $l$-th hidden layer of an $L$-layered DNN, $I(X;T_{l})$ gives a quantitative measure of the information in $T_{l}$.
\cite{goldfeld2018estimating} showed that this reduction in $I(X;T_{l})$ over the course of training is driven by progressive geometric clustering of the representations of samples from the same class.
The clusters tighten as we move into deeper layers providing evidence that this layered structure progressively improves representation of $X$ to increase its relevance for $Y$.
The rise and fall of $I(X;T_{l})$ corresponds to how spread out or clustered the representations in each layer are.

% Need to introduce Goldfeld's differential estimator because clustering is related to fluctuations in I(X;T_{l}) in low-beta regime
 \cite{goldfeld2018entropy} proposes a {\it noisy DNN framework} to estimate $I(X;T_{l})$. In this framework, each neuron adds a small amount of Gaussian noise $\mathcal{Z} \sim \mathcal{N}(0,\beta^{2}I_{d})$ (iid across neurons) after applying the activation functions, where $d$ denotes the number of features in the $l-1$-th layer representation.
Although injection of this noise renders $I(X;T_{l})$ meaningful for studying deep learning, the concatenation of Gaussian noises and non-linearities makes the mutual information impossible to compute analytically or evaluate numerically.
\cite{goldfeld2018estimating} uses this noisy DNN framework to estimate $I(X;T_{l})$.
Smaller $\beta$ values correspond to narrow Gaussians while larger $\beta$ values correspond to wider Gaussians.
When $\beta$ is small, even Gaussians that belong to the same cluster are distinguishable as long as they are not too close.
When the clusters tighten, this in-class movement brings these Gaussians closer together effectively merging them. This causes a reduction in $I(X;T_{l})$.
For larger values of $\beta$, the in-class movement is blurred at the outset (before clusters tighten).
Therefore, the only effect on mutual information is the separation between the clusters. As the blobs move away from each other, the mutual information rises.
While ``clustering Gaussians" and ``decrease in mutual information" are strongly related in the low-$\beta$ regime, once the noise becomes large, these phenomena decouple, i.e, networks may cluster inputs and neurons may saturate, but this will not be reflected in decrease of mutual information.

% fall in MI <=> rise in accuracy
\cite{shwartz2017opening} argues that decrease in $I(X;T_{l})$ indicates an increase in prediction accuracy.
% fall in MI <=> rise in clustering of samples of the same class
We have seen that reduction in $I(X;T_{l})$ over the course of training is driven by progressive geometric clustering of the representations of samples from the same class.
% therefore, rise in accuracy <=> rise in clustering of samples of the same class
Therefore, an increase in prediction accuracy is driven by this phenomenon of progressive geometric clustering.

\subsection{Restricted Boltzmann Machines} \label{sec:RBM}

Restricted Boltzmann machines (RBMs) \cite{goodfellow2016deep} are a powerful class of energy-based models (EBMs). These are called EBMs because each configuration that the RBM can be associated with has a corresponding energy which indicate a point in the resulting energy contour. If we consider an RBM with discrete visible and discrete hidden units, their joint probability distribution can be written as

\begin{align}
    P(v,h;\Theta) = \frac{\exp{(-E(v,h;\Theta))}}{Z(\Theta)}
\end{align}

where the partition function, $Z(\Theta)$ is given by

\begin{align}
    Z(\Theta) = \sum\limits_{v} \sum\limits_{h} \exp{(-E(v,h;\Theta))}
\end{align}

and the energy function is defined as

\begin{align}
    E(v,h;\Theta) = -b_{v}^{T}\cdot v - v^{T}\cdot W\cdot h - b_{h}^{T}\cdot h
\end{align}

\paragraph{Content addressable memory}
An RBM can be considered as a {\it content addressable memory}~\cite{nagatani2014restricted}. 
By a content-addressable memory system, we mean that an RBM is designed to store a number of patterns so that they can be retrieved from noisy or partial cues. 
It does this by creating an energy surface which has minima representing each of the patterns. The noisy and partial cues are states of the system which are close to these minima. 
As an RBM evolves, it slides from the noisy pattern down the energy surface into the closest minima - representing the closest stored pattern. 
For example, train a DNN on a set of images. Then present this network with either a portion of one of the images (partial cue) or an image degraded with noise (noisy cue), sampling from the system will attempt to reconstruct one of the stored images. 

When dealing with real-valued inputs, we commonly use a {\it Gaussian-Bernoulli} RBM (GB-RBM), which has Gaussian-distributed visible units and Bernoulli-distributed hidden units. 
With $v \in \mathbb{R}^{|V|}$, $h \in \{0,1\}^{|H|}$ and $\Theta = (b_{v}, \Sigma, W, b_{h})$, the energy function becomes

\begin{align} \label{eqn:GB-RBM_E}
    E(v,h;\Theta) &= \sum\limits_{i} \frac{(v_{i}-b_{v_{i}})^{2}}{2\sigma_{i}^{2}} - \sum\limits_{ij} \frac{v_{i}}{\sigma_{i}} W_{ij} h_{j} - \sum\limits_{j} b_{h_{j}} h_{j} \nonumber \\
    &= \frac{1}{2} \left( \frac{v-b_{v}}{\sigma} \right)^{T} \cdot \left( \frac{v-b_{v}}{\sigma} \right) - \left( \frac{v}{\sigma} \right)^{T} \cdot W \cdot h - b_{h}^{T} \cdot h
\end{align}

The conditional distributions can be written as

\begin{align} \label{eqn:GB-RBM_vh}
    P(v_{i}=x|h;\Theta) &= \frac{1}{\sqrt{2\pi} \sigma_{i}}\exp \left( -\frac{1}{2\sigma_{i}^{2}}(v-b_{v_{i}}-\sigma_{i}\sum\limits_{j}W_{ij}h_{j})^{2} \right) \nonumber \\
    &= \mathcal{N}(v_{i};b_{v_{i}} + \sigma_{i}\sum\limits_{j}W_{ij}h_{j},\sigma_{i}^{2})
\end{align}

\begin{align} \label{eqn:GB-RBM_hv}
    P(h_j=1|v;\Theta) &= g(b_{h_{j}} + \sum\limits_{i}\frac{v_{i}}{\sigma_{i}}W_{ij})
\end{align}

where $g$ stands for any non-linearity function.

We encourage the reader to refer~\cite{melchior2012learning} for a detailed analysis of RBMs, particularly GB-RBMs and their applications on natural image datasets.
%%%%%%%%%%%%%%%%%%%%%%%%%%%%%%%%%%%%%%%%%%%%%%%%%%%%%%%%%%%%%%
% Motivation
%%%%%%%%%%%%%%%%%%%%%%%%%%%%%%%%%%%%%%%%%%%%%%%%%%%%%%%%%%%%%%
\section{Motivation}\label{sec:motive}
\paragraph{Problem statement} Given clean training dataset and noisy test dataset for a prediction task, our goal is to denoise the test data.

% Why use RBM?
Section~\ref{sec:RBM} showed that an RBM trained to store patterns can be used to denoise a distorted version of any of these patterns. 
Therefore, if we train an RBM on a training set and posed it with distorted version of the test set, sampling from the RBM should return a denoised version of this test set. 

% What happens with graph-structured dataset?
If the dataset is graph-structured then we are potentially dealing with noise in the node properties, edge existence and its properties.
Although, an RBM can denoise the node properties and edge properties of the test set. 
It will not be very effective for denoising the adjacency matrix because the adjacency matrix of most real-world datasets are far too sparse for any RBM to capture useful information.

% How do we deal with this adjacency matrix problem?
GNNs can learn representations that can capture both the node properties and the graph structure.
Most variants of GNNs build representations by borrowing information relevant to a node from that node's neighborhood.
Therefore, the correct neighborhood is required for a node so that the right information can be aggregated. 
If there is a way to obtain this right aggregated information, the correct neighborhood is no longer a concern.
If we train an RBM on the representations that a GNN learns from the training set, we should be able to denoise the properties of the nodes, properties of the edges as well as the adjacency matrix in the test set.

% Denoising pipeline
This leads us to a rough outline of a denoising pipeline.
\begin{enumerate}
    \item Train a DNN on the training data to obtain a concise representation of the entire information. We use a GNN when dealing with graph-structured datasets and multi-layer preceptron (MLP) for other datasets.
    \item Train an RBM on this concise representation.
    \item Pass the noisy test data through this trained DNN to obtain noisy representations.
    \item If we feed this noisy representation through our trained RBM, the subsequent reconstruction must be a denoised version of this noisy test data.
\end{enumerate}

% Which layer to use?
However, this brings up an interesting question. 
A MLP-based DNN usually contains several layers while a GNN-based one usually contains two or three hidden layers. Which of these hidden layer representations must we use for our denoising task? 
This is the question the paper attempts to answer.
%%%%%%%%%%%%%%%%%%%%%%%%%%%%%%%%%%%%%%%%%%%%%%%%%%%%%%%%%%%%%%
% Denoising pipeline
%%%%%%%%%%%%%%%%%%%%%%%%%%%%%%%%%%%%%%%%%%%%%%%%%%%%%%%%%%%%%%
\section{Denoising pipeline}\label{sec:denoise}
In this section, we detail our proposed pipeline illustrated in figure~\ref{fig:denoise} for the denoising task.

% Denoising pipeline figure
\tikzstyle{data} = [rectangle, minimum width=1cm, minimum height=1cm, text centered, draw=black, fill=blue!30]
\tikzstyle{NN} = [rectangle, minimum width=2cm, minimum height=1cm, text centered, draw=black, fill=orange!30]
\tikzstyle{post} = [rectangle, minimum width=2cm, minimum height=1cm, text centered, draw=black, fill=brown!30]
\tikzstyle{RBM} = [rectangle, minimum width=2cm, minimum height=1cm, text centered, draw=black, fill=green!30]
\tikzstyle{decision} = [diamond, minimum width=2cm, minimum height=1cm, text centered, draw=black, fill=red!30]
\tikzstyle{arrow} = [thick,->,>=stealth]

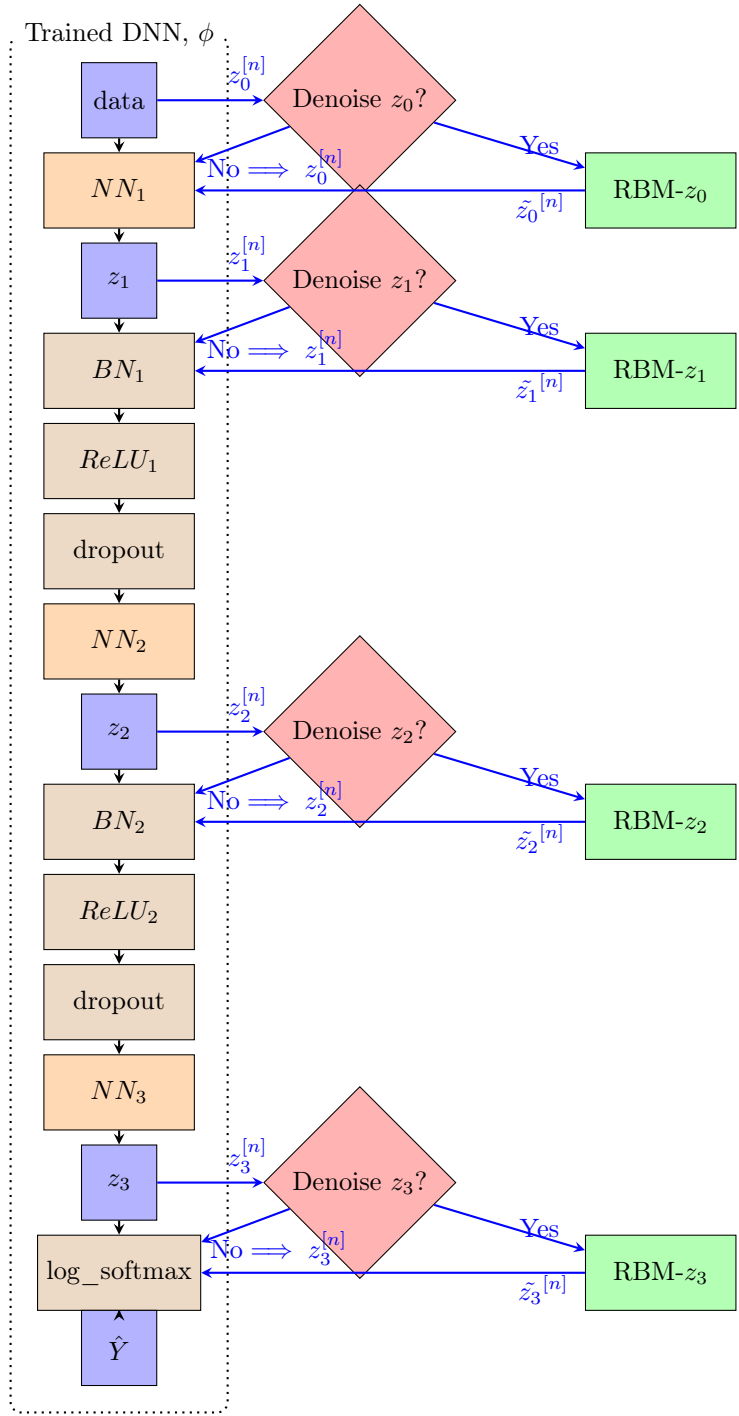
\begin{figure}[htpb]
	\centering
	\begin{tikzpicture}[node distance=1.2cm]
		%%%%%%%%%%%%%%%%%%%%%%%%%%%%
		% graph ML pipeline
		%%%%%%%%%%%%%%%%%%%%%%%%%%%%
		% nodes
		\node (x) [data] {data};
		
		\node (nn_1) [NN, below of=x] {$NN_1$};
		\node (z1) [data, below of=nn_1] {$z_{1}$};
		\node (bn_1) [post, below of=z1] {$BN_1$};
		\node (relu_1) [post, below of=bn_1] {$ReLU_1$};
		\node (dropout_1) [post, below of=relu_1] {dropout};
		
		\node (nn_2) [NN, below of=dropout_1] {$NN_2$};
		\node (z2) [data, below of=nn_2] {$z_{2}$};
		\node (bn_2) [post, below of=z2] {$BN_2$};
		\node (relu_2) [post, below of=bn_2] {$ReLU_2$};
		\node (dropout_2) [post, below of=relu_2] {dropout};
		
		\node (nn_3) [NN, below of=dropout_2] {$NN_3$};
		\node (z3) [data, below of=nn_3] {$z_{3}$};
		\node (ls) [post, below of=z3] {log\_softmax};
		
		\node (yhat) [data, below of=ls, yshift=0.2cm] {$\hat{Y}$};
		% edges
		\draw [arrow] (x) -- (nn_1);
		\draw [arrow] (nn_1) -- (z1);
		\draw [arrow] (z1) -- (bn_1);
		\draw [arrow] (bn_1) -- (relu_1);
		\draw [arrow] (relu_1) -- (dropout_1);
		\draw [arrow] (dropout_1) -- (nn_2);
		\draw [arrow] (nn_2) -- (z2);
		\draw [arrow] (z2) -- (bn_2);
		\draw [arrow] (bn_2) -- (relu_2);
		\draw [arrow] (relu_2) -- (dropout_2);
		\draw [arrow] (dropout_2) -- (nn_3);
		\draw [arrow] (nn_3) -- (z3);
		\draw [arrow] (z3) -- (ls);
		\draw [arrow] (ls) -- (yhat);
		% outline
		\node[draw, thick, dotted, rounded corners, inner xsep=1em, inner ysep=1em, fit=(x) (nn_1) (z1) (bn_1) (relu_1) (dropout_1) (nn_2) (z2) (bn_2) (relu_2) (dropout_2) (nn_3) (z3) (ls) (yhat)] (box) {};
		\node[fill=white] at (box.north) {Trained DNN, $\phi$};
		%%%%%%%%%%%%%%%%%%%%%%%%%%%%%%%%%%%%%%
		% denoising pipeline
		%%%%%%%%%%%%%%%%%%%%%%%%%%%%%%%%%%%%%%
		% nodes
		\node (denoise_z0) [decision, right of=x, xshift=2cm] {Denoise $z_{0}$?};
		\node (rbm_z0) [RBM, right of=nn_1, xshift=6cm] {RBM-$z_{0}$};
		
		\node (denoise_z1) [decision, right of=z1, xshift=2cm] {Denoise $z_{1}$?};
		\node (rbm_z1) [RBM, right of=bn_1, xshift=6cm] {RBM-$z_{1}$};
		
		\node (denoise_z2) [decision, right of=z2, xshift=2cm] {Denoise $z_{2}$?};
		\node (rbm_z2) [RBM, right of=bn_2, xshift=6cm] {RBM-$z_{2}$};
		
		\node (denoise_z3) [decision, right of=z3, xshift=2cm] {Denoise $z_{3}$?};
		\node (rbm_z3) [RBM, right of=ls, xshift=6cm] {RBM-$z_{3}$};
		% edges
		\draw [arrow, color=blue] (denoise_z0) -- node [anchor=west] {Yes} (rbm_z0);
		\draw [arrow, color=blue] (denoise_z1) -- node[anchor=west] {Yes} (rbm_z1);
		\draw [arrow, color=blue] (denoise_z2) -- node[anchor=west] {Yes} (rbm_z2);
		\draw [arrow, color=blue] (denoise_z3) -- node[anchor=west] {Yes} (rbm_z3);
		% outline
		%\node[draw, thick, dotted, rounded corners, red, inner xsep=2em, inner ysep=2em, fit=(x) (nn_1) (z1) (bn_1) (relu_1) (dropout_1) (nn_2) (z2) (bn_2) (relu_2) (dropout_2) (nn_3) (z3) (ls) (yhat) (denoise_z0) (rbm_z0) (denoise_z1) (rbm_z1) (denoise_z2) (rbm_z2)  (denoise_z3) (rbm_z3)] (box) {};
		%\node[fill=white] at (box.north) {Denoising pipeline, $\psi$};
		%%%%%%%%%%%%%%%%%%%%%%%%%%%%%%%%%%%%%%
		% inference time
		%%%%%%%%%%%%%%%%%%%%%%%%%%%%%%%%%%%%%%
		% edges
		\draw [arrow, color=blue] (x) -- node[anchor=south, xshift=0.5cm] {$z_{0}^{[n]}$} (denoise_z0);
		\draw [arrow, color=blue] (denoise_z0) -- node[anchor=west, xshift=-0.6cm, yshift=-0.3cm] {No$\implies z_{0}^{[n]}$} (nn_1);
		\draw [arrow, color=blue] (rbm_z0) -- node[anchor=north, color=blue, xshift=2cm, yshift=0.1cm] {$\tilde{z_{0}}^{[n]}$} (nn_1);
		
		\draw [arrow, color=blue] (z1) -- node[anchor=south, xshift=0.5cm] {$z_{1}^{[n]}$} (denoise_z1);
		\draw [arrow, color=blue] (denoise_z1) -- node[anchor=west, xshift=-0.6cm, yshift=-0.3cm] {No$\implies z_{1}^{[n]}$} (bn_1);
		\draw [arrow, color=blue] (rbm_z1) -- node[anchor=north, color=blue, xshift=2cm, yshift=0.1cm] {$\tilde{z_{1}}^{[n]}$} (bn_1);
		
		\draw [arrow, color=blue] (z2) -- node[anchor=south, xshift=0.5cm] {$z_{2}^{[n]}$} (denoise_z2);
		\draw [arrow, color=blue] (denoise_z2) -- node[anchor=west, xshift=-0.6cm, yshift=-0.3cm] {No$\implies z_{2}^{[n]}$} (bn_2);
		\draw [arrow, color=blue] (rbm_z2) -- node[anchor=north, color=blue, xshift=2cm, yshift=0.1cm] {$\tilde{z_{2}}^{[n]}$} (bn_2);
		
		\draw [arrow, color=blue] (z3) -- node[anchor=south, xshift=0.5cm] {$z_{3}^{[n]}$} (denoise_z3);
		\draw [arrow, color=blue] (denoise_z3) -- node[anchor=west, xshift=-0.6cm, yshift=-0.3cm] {No$\implies z_{3}^{[n]}$} (ls);
		\draw [arrow, color=blue] (rbm_z3) -- node[anchor=north, color=blue, xshift=2cm, yshift=0.1cm] {$\tilde{z_{3}}^{[n]}$} (ls);
	\end{tikzpicture}
	\caption{Denoising pipeline}
	\label{fig:denoise}
\end{figure}

\subsection{Trained deep neural network, $\phi$}~\label{sec:phi}
$\psi$ denotes the DNN that performs the downstream ML task.
As a pedagogical tool, we depict a DNN with three hidden layers $NN_{1}$, $NN_{2}$ and $NN_{3}$, where $NN$ is simply a placeholder for any neural network layer. 
{\it data} denotes the graph input. 
The hidden layers representations after training are denoted by $z_{1}$, $z_{2}$ and $z_{3}$ respectively. 
We apply batch normalization ({\it BN}), followed by ReLU non-linearity ({\it relu}) and then dropout regularization ({\it dropout}) for each hidden layer representation before feeding it to the subsequent hidden layer. 
The predicted output $\hat{Y}$ is the logarithm of the softmax of $z_{3}$. 

We will denote the DNN containing NN layers which is trained on our clean training set by $\phi^{NN}$. 
$z_{0}^{[n]}$ will interchangeably be used alongside $data^{[n]}$ to indicate test data with $n\%$ noise.
$\phi^{NN}(data^{[n]})$ indicates when $data^{[n]}$ is fed to the trained pipeline $\phi^{NN}$. 
$\phi^{NN}(data^{[n]})$ gives us $z_{1}^{[n]}$, $z_{2}^{[n]}$ and $z_{3}^{[n]}$ which are the noisy estimates at the first, second and third hidden layers respectively.

\subsection{Denoising pipeline, $\psi$}~\label{sec:psi}
Here, we describe the pipeline which is responsible for denoising the test data based on representations $z_{i}^{[n]}, i = \{0,1,2,3\}$.
First we pick a layer $i$ whose embeddings we consider are most informative for our denoising task. 
We use {\it RBM-$z_{i}$} to denote the RBM trained on $z_{i}$.
Since $z_{i}^{[n]}$ is a noisy estimate of some true representation that we would get if the data was noise-free, we feed these to {\it RBM-$z_{i}$} and obtain a sample $\tilde{z_{i}}^{[n]}$. 
These reconstructions should have evolved towards the closest minimum in the energy surface learned by {\it RBM-$z_{i}$}, and therefore we are closer to the true representations. 
When $\tilde{z_{i}}^{[n]}$ is passed back into the DNN at the layer succeeding $z_{i}$, and then progressed through the rest of the pipeline, we should get the true prediction $\hat{Y}$.
%%%%%%%%%%%%%%%%%%%%%%%%%%%%%%%%%%%%%%%%%%%%%%%%%%%%%%%%%%%%%%
% Experiments
%%%%%%%%%%%%%%%%%%%%%%%%%%%%%%%%%%%%%%%%%%%%%%%%%%%%%%%%%%%%%%
\section{Experiments}\label{sec:expts}
We now evaluate the effectiveness of our denoising pipelines on the task of node property prediction.
We train on the original node feature matrix and the adjacency matrix corresponding to the dataset, but with synthetic noisy validation and test sets.

\subsection{Dataset}
We use the following two citation datasets:
\begin{itemize}
    \item {\bf ogbn-arxiv} paper citation network to demonstrate denoising in near-realistic situations.
    \item {\bf WikiCS} article citation network for ease in analyzing progressive geometric clustering which is described in section~\ref{sec:discuss}.
\end{itemize}, 

\subsubsection{ogbn-arxiv paper citation network}\label{sec:ogbn-arxiv}
% What is OGB?
The {\it Open Graph Benchmark} (OGB)~\cite{hu2020open} is a diverse set of challenging and realistic benchmark datasets to facilitate scalable, robust, and reproducible graph ML research. 
% Why OGB?
We chose OGB for the following reason:
The commonly used node classification datasets are {\it Cora}, {\it Citeseer} and {\it Pubmed} which only have $2700$ to $20000$ nodes~\cite{yang2016revisiting}. 
Such small datasets make it hard to rigorously evaluate data-hungry deep learning models. 
The performance of GNNs on these datasets is often unstable and nearly statistically identical to each other, due to the small number of samples the models are trained and evaluated on.
Furthermore, different studies adopt their own dataset splits, evaluation metrics, and cross-validation protocols making it harder to compare performance across various studies~\cite{dwivedi2020benchmarking}.
Additionally, ogbn-arxiv provides a unified evaluation protocol using meaningful application-specific data splits and evaluation metrics, accompanied by extensive benchmark experiments for each dataset. 
What makes OGB of particular interest is that it provides an automated end-to-end graph ML pipeline that simplifies and standardizes the process of graph data loading, experimental setup, and model evaluation.

% What is ogbn-arxiv and why choose this?
The {\it ogbn-arxiv} dataset is a directed graph, denoting the citation network between all Computer Science (CS) arXiv papers extracted from the Microsoft Academic Graph (MAG)~\cite{wang2019deep}. 
% Description for the nodes and the edges 
Each node is an arXiv paper and each directed edge indicates that one paper cites another one. There are $169343$ nodes and $1166243$ edges making them large enough for rigorous graph ML applications but small enough to fit into the memory of a single GPU.
This is the reason behind choosing this dataset from all the available ones in~\cite{hu2020open}.
% Description for the node feature matrix
Each paper comes with a $128$ dimensional feature vector obtained by averaging the embeddings of words in its title and abstract. The embeddings of individual words are computed by running the {\it word2vec} model~\cite{mikolov2013distributed} over the MAG corpus. 
All papers are also associated with the year that the corresponding paper was published.

% Dataset splits
To counter the problematic arbitrary spliting of datapoints, the {\it ogbn-arxiv} dataset uses a realistic data split based on the publication dates of the papers. The training set contains papers published until 2017, papers published in 2018 form the validation set, and the testing set is composed of papers published since 2019. 

% Original node property prediction task
The task is to predict the primary categories of the arXiv papers into one of the 40 subject areas of arXiv CS papers.

% Baseline models for ogbn-arxiv
\cite{hu2020open} provides the following models as baselines where each model has $256$ hidden units, $2$ hidden layers where each layer has a dropout of $0.5$.
\begin{itemize}
    \item {\bf MLP}: A multi-layer perceptron (MLP) that uses the raw node features directly as input without accounting for the graph structure.
    \item {\bf node2vec}: An MLP predictor that uses as input the concatenation of the raw node features and {\it node2vec} embeddings~\cite{grover2016node2vec}.
    \item {\bf GCN}: Full-batch Graph Convolutional Network~\cite{kipf2016semi}.
    \item {\bf GraphSAGE}: Full-batch GraphSAGE~\cite{hamilton2017inductive} where mean pooling variant and simple skip connection is adopted to preserve central node features.
\end{itemize}

\subsubsection{WikiCS article citation network}\label{sec:wikics}
% What is WikiCS?
{\it WikiCS} article citation network~\cite{mernyei2020wiki} is a dataset consisting of nodes corresponding to CS articles with edges based on hyperlinks and $10$ classes representing different branches of the field. This dataset is primarily focused on semi-supervised node classification: given the labels are provided for a small fraction of the nodes (typically 1-5\%), features for all nodes, and their connectivity, the task is to predict all other labels.
% Description for the nodes, node feature matrix and the edges
The dataset consists of $11701$ nodes and $216123$ edges. Similar to the {\it ogbn-arxiv} dataset, each node embedding was derived for the text of the corresponding articles. Here the embeddings were calculated as the average of pretrained GloVe word embeddings~\cite{pennington2014glove}. This condenses the information present in the articles into a $300$ dimensional input vector. 

% Dataset splits
The dataset offers $20$ different training splits from the data that was not used for testing. The nodes in all the splits are distributed as follows: $580$ nodes for the training set, $1769$ nodes for the validation set and $5847$ nodes for the test set. 
For our experiments, we make the following two changes: (1) we only make use of the first split, (2) we place all the nodes in this split that do not belong to the training, validation or test sets to the training set because the original semi-supervised learning setup of the dataset is not of interest to us. 
Thus, our training set now contains $4085$ nodes.

% Original node property prediction task
Similar to section~\ref{sec:ogbn-arxiv}, the task is to classify the articles into one of the 10 areas of CS.

% Why WikiCS?
We chose WikiCS dataset particularly for the ease in viewing the progressive geometric clustering of representations as described in section~\ref{sec:bg}. Since this validation set consists of only $1769$ nodes and the test set contains $5847$ nodes, this makes viewing the internal representation space of the denoising pipeline rather easy.

% Baseline models
We use the same baselines as described in section~\ref{sec:ogbn-arxiv} with each model having $35$ hidden units, $2$ hidden layers where each layer has a dropout with probability $0.35$, and learning rate of $0.003$.

\subsubsection{Synthetic noisy data}
For both {\it ogbn-arxiv} and {\it WikiCS}, we will assume that we are given a combination of (1,3), (1,4), (2,3) or (2,4) from the following listed possibilities:

\begin{enumerate}
    \item {\it Corrupted node feature matrix}, $X_{c}$: This will occur when the title and the abstract contain words that are confusing or misleading. These words may be chosen, intentionally or otherwise, to showcase the paper as something it is not necessarily.
    \item {\it Partial node feature matrix}, $X_{z}$: If the title or abstract is very sparingly or vaguely worded, the generated node feature vectors may contain a lot of zeroes.
    \item {\it Corrupted adjacency matrix}, $A_{c}$: Due to tricks used by journals to boost their impact factor or for other reasons, a paper might have links to others that are not really necessary.
    \item {\it Partial adjacency matrix}, $A_{z}$: This will occur when paper has too few citations.
\end{enumerate}

We maintain the same node property prediction task as was assigned for the original dataset. But we have the added constraint of having to resolve the noisy data before proceeding with this prediction task.

 For $n = \{0, 10, \dots, 100\}$, where $n$ is the percentage of distortion, we prepare the noisy sets as:

\begin{enumerate}
    \item $X_{c}^{[n]}$: We corrupt the node feature vectors by adding a uniform random number between $0$ and $1$ to $n\%$ of entries in the original validation and test node feature vectors.
    \item $X_{z}^{[n]}$: For the partial node feature vectors, we generate them by blanking out $n\%$ of entries in the original validation and test node feature vectors.
    \item $A_{c}^{[n]}$: For $n\%$ of the nodes, the adjacency matrix is corrupted by replacing the true neighbour of a node in the validation set with a random node in the training set for {\it ogbn-arxiv}, and with a random node from any of the training, validation or test sets in the case of {\it WikiCS}.
    
    For the nodes in the test set, we replace the true neighbour with a node in either the training set or the validation set for {\it ogbn-arxiv}, and with any random node from the training, validation or the test sets for {\it WikiCS}.
    
    This is because the {\it ogbn-arxiv} is split by publication date while such data is not available in the case of {\it WikiCS}.
    \item $A_{z}^{[n]}$: Blanking out in the adjacency matrix is achieved by eliminating $n\%$ of edges.
\end{enumerate}

We will use $X$ when we have to refer to both $X_{c}$ and $X_{z}$. The same approach is followed when using $A$.

Our experiments were performed on a Nvidia Tesla T4 GPU with 16GB memory.
%%%%%%%%%%%%%%%%%%%%%%%%%%%%%%%%%%%%%%%%%%%%%%%%%%%%%%%%%%%%%%
% Results
%%%%%%%%%%%%%%%%%%%%%%%%%%%%%%%%%%%%%%%%%%%%%%%%%%%%%%%%%%%%%%
\section{Results}\label{sec:results}
We use $\mathcal{P}$ to denote the prediction accuracy of any of these pipelines. 
If $n_{X}$ and $n_{A}$ denote the percentage of noise in the node feature matrix and the adjacency matrix respectively, we will compare $\mathcal{P}(\phi^{NN}(X^{[n_{X}]},A^{[n_{A}]}))$, 
$\mathcal{P}(\psi_{0}^{NN}(X^{[n_{X}]},A^{[n_{A}]}))$, 
$\mathcal{P}(\psi_{1}^{NN}(X^{[n_{X}]},A^{[n_{A}]}))$, 
$\mathcal{P}(\psi_{2}^{NN}(X^{[n_{X}]},A^{[n_{A}]}))$ and \\
$\mathcal{P}(\psi_{3}^{NN}(X^{[n_{X}]},A^{[n_{A}]}))$ for each of the above mentioned baselines and for $n_{X}={\{0,10,\dots,100\}}, n_{A}=\{0,10,\dots,100\}$.

We have observed that RBMs with 4096 hidden units are the most efficient for learning the representations when each of these are trained for $1000$ epochs where datapoints are in batches of size $64$ with $1$ step of contrastive divergence.

Due to space limitations and for efficient viewing of the results, we have hosted the results as interactive web applications which will be explained in this section. 
The reader is encouraged to load the web application alongside this paper for ease in understanding.

\subsection{Prediction accuracy}
The prediction accuracy results are hosted at \url{https://ankithmo.shinyapps.io/denoiseRBM}.
This application is explained as follows:

\paragraph{Dashboard}
The user is provided with the following controls:
(1) dataset (ogbn-arxiv or WikiCS), 
(2) split (validation or test set), 
(3) type of distortion in node feature matrix $X$ (corrupted $X_{c}$ or blanked out $X_{z}$), 
(4) type of distortion in adjacency matrix $A$ (corrupted $A_{c}$ or blanked out $A_{z}$), 
(5) amount of distortion in adjacency matrix $n_{A}$ ($0\%$ to $100\%$ in increments of $10\%$), and 
(6) various reconstructions to be used for comparison. 

\paragraph{Accuracy} 
$\mathcal{P}(\psi_{i}^{NN}(X^{[n_{X}]},A^{[n_{A}]})), i=\{0,1,2,3\}$ and $\mathcal{P}(\phi^{NN}(X^{[n_{X}]},A^{[n_{A}]}))$ where $n_{X} \in {0,10,\dots,100}$ are plotted.
A separate tab exists for each \\ $NN \in \{MLP, n2v, GCN, SAGE\}$.

\paragraph{Playground}
This compares $\phi^{NN}(X^{[n_{X}]},A^{[n_{A}]})$ where $n_{X} \in {0,10,\dots,100}$.
These plots exist by default and the user can remove any or all of these.
Also the user is allowed to add $\psi_{i}^{NN}(X^{[n_{X}]},A^{[n_{A}]})$ for any value of $i \in \{0,1,2,3\}$ and $NN \in \{MLP, n2v, GCN, SAGE\}$.

\subsection{Progressive geometric clustering}\label{sec:cluster}
The results of tracking the geometric clustering with changes in the prediction accuracy is hosted in a separate web application for each neural network model. These are available at the following URLs:
\begin{itemize}
    \item MLP: \url{https://ankithmo.shinyapps.io/denoiseRBM_viz_MLP/}
    \item node2vec: \url{https://ankithmo.shinyapps.io/denoiseRBM_viz_n2v/}
    \item GCN: \url{https://ankithmo.shinyapps.io/denoiseRBM_viz_GCN/}
    \item SAGE: \url{https://ankithmo.shinyapps.io/denoiseRBM_viz_SAGE/}
\end{itemize}
These applications follow a common template which is explained as follows:

\paragraph{Dashboard} 
The user has the following options: 
(1) type of distortion in adjacency matrix $A$ (corrupted $A_{c}$ or blanked out $A_{z}$), 
(2) amount of distortion in adjacency matrix $n_{A}$ ($0\%$ to $100\%$ in increments of $10\%$), 
(3) type of distortion in node feature matrix $X$ (corrupted $X_{c}$ or blanked out $X_{z}$), and 
(4) amount of distortion in node feature matrix $n_{X}$ ($0\%$ to $100\%$ in increments of $10\%$).
In the case of MLP, since distortions in the adjacency matrix do not make any impact, options (3) and (4) are not provided.

\paragraph{Accuracy} 
$\mathcal{P}(\psi_{i}^{NN}(X^{[n_{X}]},A^{[n_{A}]})), i=\{0,1,2,3\}$ and $\mathcal{P}(\phi^{NN}(X^{[n_{X}]},A^{[n_{A}]}))$ \\ where $n_{X} \in {0,10,\dots,100}$ are plotted.

\paragraph{Desired representations} 
$z_{i}, i=\{0,1,2,3\}$ are the t-SNE~\cite{maaten2008visualizing} representations of the clean validation/test set. 
These are the embeddings that we expect the denoising pipeline to generate.

\paragraph{Denoised data representations}
$NN:\sim z_{i}\ representations, i=\{0,1,2,3\}$ shows the t-SNE embeddings of the input layer and the hidden layer representations when the layer following $z_{i}$ in the denoising pipeline is presented with the reconstructions generated by RBM-$z_{i}$.

\paragraph{Noisy data representations}
$NN:z0~z_{i}, i=\{0,1,2,3\}$ which correspond to the t-SNE embeddings of the input layer and the hidden layer representations when the denoising pipeline is presented with noisy data.

To prevent the plot from getting cluttered with text, we use $NN$ to denote $\mathcal{P}(\phi^{NN}(X^{[n]},A^{[n]}))$, $NN:~x$ for $\mathcal{P}(\psi_{0}^{NN}(X^{[n]},A^{[n]}))$, $NN:~z1$ for $\mathcal{P}(\psi_{1}^{NN}(X^{[n]},A^{[n]}))$, $NN:~z2$ for $\mathcal{P}(\psi_{2}^{NN}(X^{[n]},A^{[n]}))$ and $NN:~z3$ to indicate
$\mathcal{P}(\psi_{3}^{NN}(X^{[n]},A^{[n]}))$.
%%%%%%%%%%%%%%%%%%%%%%%%%%%%%%%%%%%%%%%%%%%%%%%%%%%%%%%%%%%%%%
% Analysis
%%%%%%%%%%%%%%%%%%%%%%%%%%%%%%%%%%%%%%%%%%%%%%%%%%%%%%%%%%%%%%
\section{Analysis}\label{sec:analysis}
% Only important observations
In this section, we will restrict our focus to the important observations.

% Organization of analysis section
We organize this section in the following manner. 
First we will look at typical observations in traditional NN-based denoising pipelines (tNN) which will be followed by GNN-based ones (GNN).
We have illustrated plots representative of the performance of the different layers under the various distortions for ease in referring to them.
In some circumstances, we will highlight some unexpected behaviour exhibited by the pipelines.
Due to space limitations, we restrict this section to focus on the plots that are representative of the behaviour exhibited by the denoising pipelines.
We believe that these plots are sufficient for the overall understanding of the behaviour of the pipelines.
The other plots concerning some interesting atypical behaviour are presented in the appendix. 
%%%%%%%%%%%%%%%%%%%%%%%%%%%%%%%%%%%%%%%%%%%%%%%%%%%%%%%%%%%%%%
% Traditional NN-based denoising pipelines
%%%%%%%%%%%%%%%%%%%%%%%%%%%%%%%%%%%%%%%%%%%%%%%%%%%%%%%%%%%%%%
% MLP- and n2v-based denoising pipelines
\subsection{Traditional NN-based denoising pipelines}\label{sec:tNN}
We will look at denoising pipelines that use MLP as the hidden layers.
MLP-based denoising pipelines use the node feature vectors completely disregarding the adjacency matrix.
node2vec-based denoising pipelines use a concatenation of node feature vectors and node2vec embeddings as input.

% MLP
\subsubsection{MLP-based denoising pipeline}\label{sec:MLP}

\begin{figure}[htpb]
    \centering
    \begin{minipage}{0.5\textwidth}
        \centering
        \includegraphics[scale=0.25]{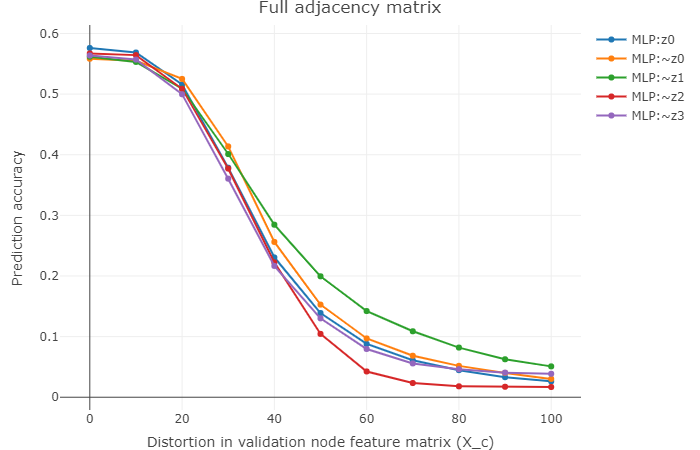}
        \subcaption{{\it ogbn-arxiv} dataset \label{fig:sub:o_MLP_X_c}}
    \end{minipage}%
    \begin{minipage}{0.5\textwidth}
        \centering
        \includegraphics[scale=0.25]{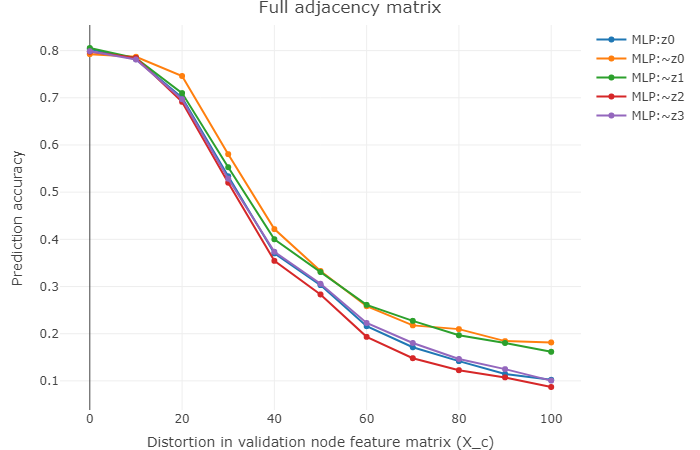}
        \subcaption{{\it WikiCS} dataset\label{fig:sub:w_MLP_X_c}}
    \end{minipage}
    \caption{$\mathcal{P}(\psi_{i}^{MLP}(X_{c}^{[n_{X}]},\cdot))$ and $\mathcal{P}(\phi^{MLP}(X_{c}^{[n_{X}]},\cdot)), i \in \{0,1,2,3\}, n_{X} \in \{0,10,\dots,100\}$.}
    \label{fig:MLP_X_c}
\end{figure}

% X_c
\paragraph{Corrupted node feature matrix ($MLP:X_{c}$)}
Figure~\ref{fig:MLP_X_c} gives us two important observations:
% obs 1 (expected)
(1) $\mathcal{P}(\psi_{1}^{MLP}(X_{c}^{[n_{X}]},\cdot))$ outperforms the rest.
In the case of the WikiCS dataset, $\mathcal{P}(\psi_{0}^{MLP}(X_{c}^{[n_{X}]},\cdot))$ gives a similar performance,
% obs 2 (Why?)
(2) $\mathcal{P}(\psi_{2}^{MLP}(X_{c}^{[n_{X}]},\cdot))$ deteriorates faster than the others.

\begin{figure}[htpb]
    \centering
    \begin{minipage}{0.5\textwidth}
        \centering
        \includegraphics[scale=0.25]{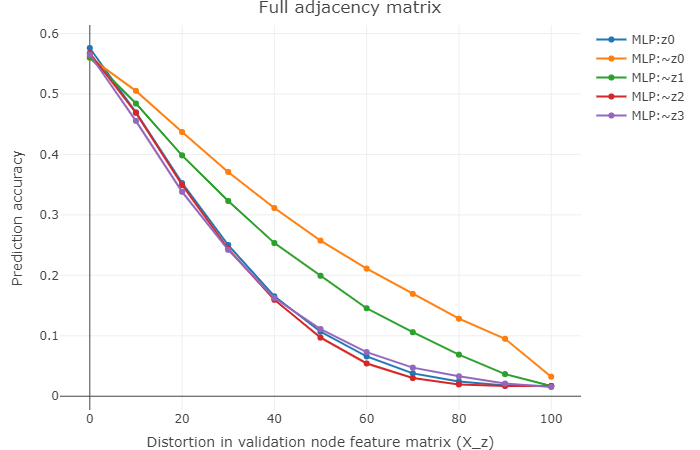}
        \subcaption{{\it ogbn-arxiv} dataset \label{fig:sub:o_MLP_X_z}}
    \end{minipage}%
    \begin{minipage}{0.5\textwidth}
        \centering
        \includegraphics[scale=0.25]{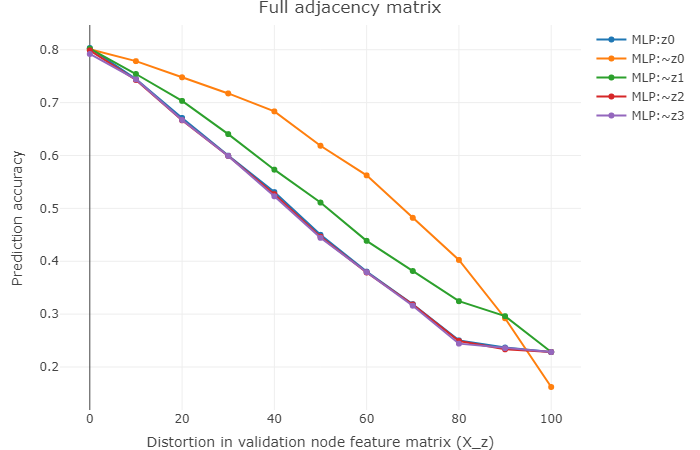}
        \subcaption{{\it WikiCS} dataset\label{fig:sub:w_MLP_X_z}}
    \end{minipage}
    \caption{$\mathcal{P}(\psi_{i}^{MLP}(X_{z}^{[n_{X}]},\cdot))$ and $\mathcal{P}(\phi^{MLP}(X_{z}^{[n_{X}]},\cdot)), i \in \{0,1,2,3\}, n_{X} \in \{0,10,\dots,100\}$.}
    \label{fig:MLP_X_z}
\end{figure}

% X_z
\paragraph{Blanked out node feature matrix ($MLP:X_{z}$)}
% obs (Why?)
We observe from figure~\ref{fig:MLP_X_z} that $\mathcal{P}(\psi_{0}^{MLP}(X_{z}^{[n_{X}]},\cdot))$ outperforms the rest and is followed by $\mathcal{P}(\psi_{1}^{MLP}(X_{z}^{[n_{X}]},\cdot))$.

% n2v
\subsubsection{node2vec-based denoising pipeline}\label{sec:n2v}
% Why no results when A_z > 40%?
Beyond $40\%$ incompletion in the adjacency matrix, we observe that the nodes in {\it ogbn-arxiv} become increasingly isolated.
node2vec embeddings cannot be computed for such nodes and therefore we omit denoising under such circumstances.

\begin{figure}[htpb]
    \centering
    \begin{minipage}{0.5\textwidth}
        \centering
        \includegraphics[scale=0.25]{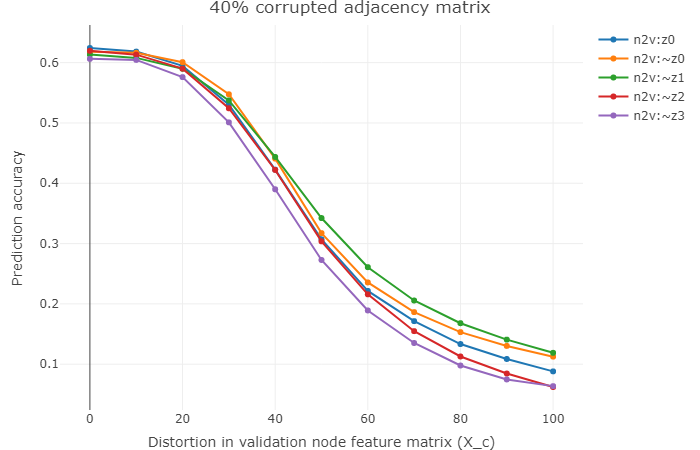}
        \subcaption{{\it ogbn-arxiv} dataset \label{fig:sub:o_n2v_X_c_A_c}}
    \end{minipage}%
    \begin{minipage}{0.5\textwidth}
        \centering
        \includegraphics[scale=0.25]{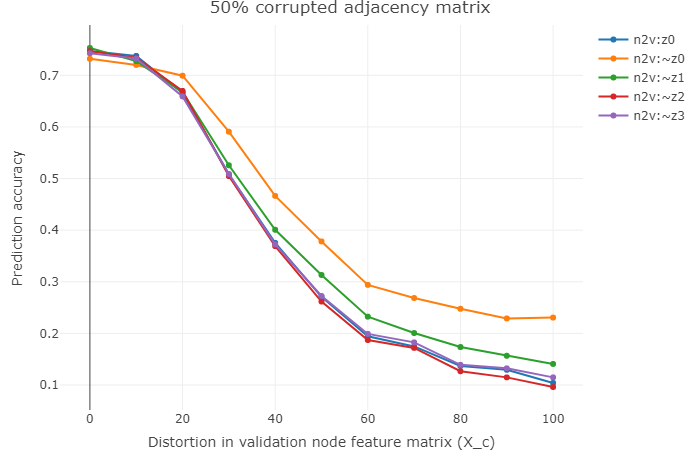}
        \subcaption{{\it WikiCS} dataset\label{fig:sub:w_n2v_X_c_A_c}}
    \end{minipage}
    \caption{Representative behaviour of $\mathcal{P}(\psi_{i}^{n2v}(X_{c}^{[n_{X}]},A_{c}^{[n_{A}]}))$ and $\mathcal{P}(\phi^{n2v}(X_{c}^{[n_{X}]},A_{c}^{[n_{A}]})), i \in \{0,1,2,3\}, n_{X} \in \{0,10,\dots,100\}, n_{A} \in \{0,10,\dots,100\}$.}
    \label{fig:n2v_X_c_A_c}
\end{figure}

% X_c, A_c
\paragraph{Corrupted node feature matrix, corrupted adjacency matrix ($n2v:X_{c},A_{c}$)}
We observe the following:
(1) Either $\mathcal{P}(\psi_{1}^{n2v}(X_{c}^{[n_{X}]},A_{c}^{[n_{A}]}))$ outperforms the rest with $\mathcal{P}(\psi_{0}^{n2v}(X_{c}^{[n_{X}]},A_{c}^{[n_{A}]}))$ closely following (see figure~\ref{fig:sub:o_n2v_X_c_A_c}), or $\mathcal{P}(\psi_{0}^{n2v}(X_{c}^{[n_{X}]},A_{c}^{[n_{A}]}))$ outperforms the rest with $\mathcal{P}(\psi_{1}^{n2v}(X_{c}^{[n_{X}]},A_{c}^{[n_{A}]}))$ following (see figure~\ref{fig:sub:w_n2v_X_c_A_c}),
(2) $\mathcal{P}(\psi_{2}^{n2v}(X_{c}^{[n_{X}]},A_{c}^{[n_{A}]}))$ shows deterioration.

% representative case figure

% Atypical case figure
Figure~\ref{fig:n2v_X_c_A_c_0} depicts a particularly interesting behaviour in WikiCS dataset where $\mathcal{P}(\psi_{i}^{n2v}(X_{c}^{[n_{X}]},A_{c}^{[n_{A}]})), i=\{1,2,3\}$ and $\mathcal{P}(\phi^{n2v}(X_{c}^{[n_{X}]},A_{c}^{[n_{A}]}))$ surpass $\mathcal{P}(\psi_{0}^{n2v}(X_{c}^{[n_{X}]},A_{c}^{[n_{A}]}))$ when $n_{X}<30$.

% X_c, A_z
\paragraph{Corrupted node feature matrix, blanked out adjacency matrix ($n2v:X_{c},A_{z}$)}
We observe from figure~\ref{fig:n2v_X_c_A_z} that: 
(1) $\mathcal{P}(\psi_{0}^{n2v}(X_{c}^{[n_{X}]},A_{z}^{[n_{A}]}))$ outperforms the rest, 
(2) This is followed by $\mathcal{P}(\psi_{1}^{n2v}(X_{c}^{[n_{X}]},A_{z}^{[n_{A}]}))$, and 
(3) $\mathcal{P}(\psi_{2}^{n2v}(X_{c}^{[n_{X}]},A_{z}^{[n_{A}]}))$ shows some deterioration but it is not significant.

\begin{figure}[htpb]
    \centering
    \begin{minipage}{0.5\textwidth}
        \centering
        \includegraphics[scale=0.25]{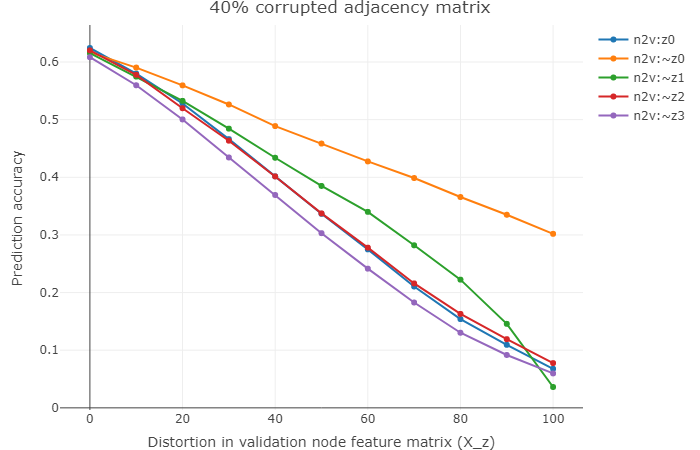}
        \subcaption{{\it ogbn-arxiv} dataset \label{fig:sub:o_n2v_X_z}}
    \end{minipage}%
    \begin{minipage}{0.5\textwidth}
        \centering
        \includegraphics[scale=0.25]{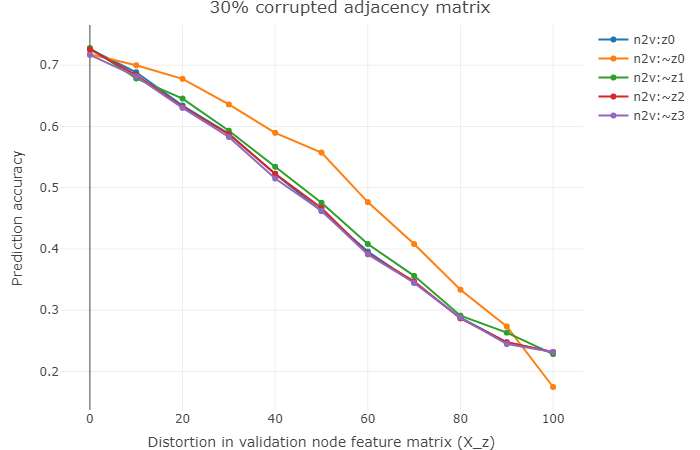}
        \subcaption{{\it WikiCS} dataset\label{fig:sub:w_n2v_X_z}}
    \end{minipage}
    \caption{Representative behaviour of $\mathcal{P}(\psi_{i}^{n2v}(X_{z}^{[n_{X}]},A^{[n_{A}]}))$ and $\mathcal{P}(\phi^{n2v}(X_{z}^{[n_{X}]},A^{[n_{A}]})), i \in \{0,1,2,3\}, n_{X} \in \{0,10,\dots,100\}, n_{A} \in \{0,10,\dots,100\}$.} 
    \label{fig:n2v_X_z}
\end{figure}

%X_z
\paragraph{Blanked out node feature matrix ($n2v:X_{z}$)}
% psi_0
For both of these datasets we observe that $\mathcal{P}(\psi_{0}^{n2v}(X_{z}^{[n_{X}]},A_{c}^{[0]}))$ exhibits a gradual decrease but manages to outperform the rest.
% the rest
\\ $\mathcal{P}(\phi^{n2v}(X_{z}^{[n_{X}]},A_{c}^{[0]}))$, $\mathcal{P}(\psi_{2}^{n2v}(X_{z}^{[n_{X}]},A_{c}^{[0]}))$ and $\mathcal{P}(\psi_{3}^{n2v}(X_{z}^{[n_{X}]},A_{c}^{[0]}))$ show a linearly decreasing trend which becomes more and more convex with increasing $n_{A}$.
% psi_1
% ogbn-arxiv
In the case of the {\it ogbn-arxiv} dataset, $\mathcal{P}(\psi_{1}^{n2v}(X_{z}^{[n_{X}]},A_{c}^{[0]}))$ exhibits a bomb-like trajectory which becomes more linear as $n_{A}$ increases. 
This trajectory 
% WikiCS
For WikiCS dataset, we observe insignificant improvement by $\mathcal{P}(\psi_{1}^{n2v}(X_{z}^{[n_{X}]},A_{c}^{[0]}))$.
This is observed in figure~\ref{fig:n2v_X_z}.

% A_z compared with A_c
When the adjacency matrix is blanked out, we observe a similar trend in performance as exhibited by corrupting the adjacency matrix.
The only notable difference is that the reduction in performance is much slower in the blanking out case when compared with the corrupted case.

% representative case figure

%%%%%%%%%%%%%%%%%%%%%%%%%%%%%%%%%%%%%%%%%%%%%%%%%%%%%%%%%%%%%%
% GNN-based denoising pipelines
%%%%%%%%%%%%%%%%%%%%%%%%%%%%%%%%%%%%%%%%%%%%%%%%%%%%%%%%%%%%%%
\subsection{GNN-based denoising pipelines}\label{sec:GNN}
Here we will cover the denoising pipelines that use either GCNs or GraphSAGE layers as the hidden layers.

% GCN
\subsubsection{GCN-based denoising pipeline}\label{sec:GCN}
When the corruption in the adjacency matrix exceeds $60\%$, the behaviour exhibited by the layers when the node feature matrix is corrupted in {\it ogbn-arxiv} dataset, and when the node feature matrix is blanked for {\it WikiCS} dataset, both lack consistency.

\begin{figure}[htpb]
    \centering
    \begin{minipage}{0.5\textwidth}
        \centering
        \includegraphics[scale=0.25]{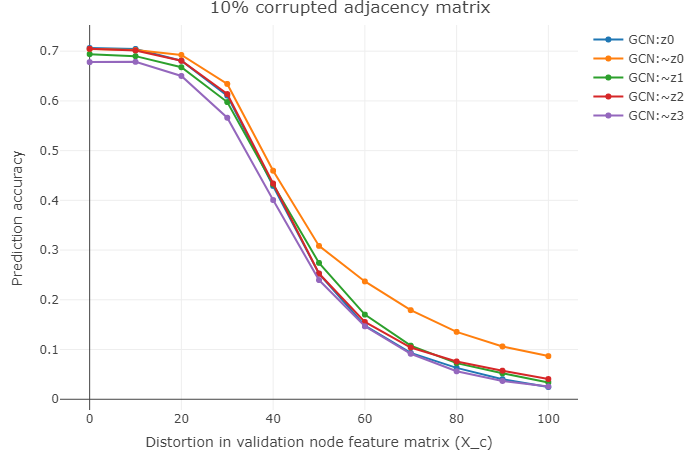}
        \subcaption{{\it ogbn-arxiv} dataset \label{fig:sub:o_GCN_X_c_A_c}}
    \end{minipage}%
    \begin{minipage}{0.5\textwidth}
        \centering
        \includegraphics[scale=0.25]{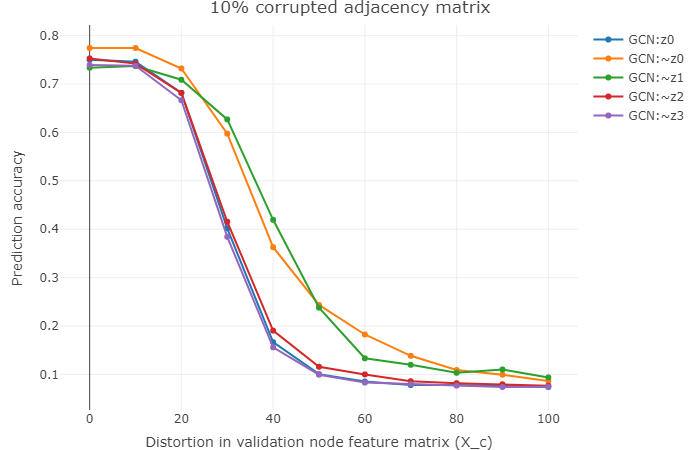}
        \subcaption{{\it WikiCS} dataset\label{fig:sub:w_GCN_X_c_A_c}}
    \end{minipage}
    \caption{Representative behaviour of $\mathcal{P}(\psi_{i}^{GCN}(X_{c}^{[n_{X}]},A_{c}^{[n_{A}]}))$ and $\mathcal{P}(\phi^{GCN}(X_{c}^{[n_{X}]},A_{c}^{[n_{A}]})), i \in \{0,1,2,3\}, n_{X} \in \{0,10,\dots,100\}, n_{A} \in \{0,10,\dots,100\}$.}
    \label{fig:GCN_X_c_A_c}
\end{figure}

% X_c
\paragraph{Corrupted node feature matrix ($GCN:X_{c}$)}
We observe the following:
% obs 1
(1) $\mathcal{P}(\psi_{0}^{GCN}(X_{c}^{[n_{X}]},A^{[n_{A}]}))$ outperforms the rest with $\mathcal{P}(\psi_{1}^{GCN}(X_{c}^{[n_{X}]},A^{[n_{A}]}))$ following (see figure~\ref{fig:GCN_X_c_A_c}),
% obs 2
(2) $\mathcal{P}(\psi_{3}^{GCN}(X_{c}^{[n_{X}]},A^{[n_{A}]}))$ usually performs the worst.

Figure~\ref{fig:GCN_X_c_A_c_0} depicts particularly interesting behaviour.
In the case of {\it ogbn-arxiv} dataset (figure~\ref{fig:sub:o_GCN_X_c_A_c_0}), $\mathcal{P}(\psi_{3}^{GCN}(X_{c}^{[n_{X}]},A_{c}^{[n_{A}]}))$ outperforms the rest including $\mathcal{P}(\phi^{GCN}(X_{c}^{[n_{X}]},A_{c}^{[n_{A}]}))$ when $n_{X}>70$.
For {\it WikiCS} dataset (figure~\ref{fig:sub:w_GCN_X_c_A_c_0}), both $\mathcal{P}(\psi_{0}^{GCN}(X_{c}^{[n_{X}]},A_{c}^{[n_{A}]}))$
 and $\mathcal{P}(\psi_{3}^{GCN}(X_{c}^{[n_{X}]},A_{c}^{[n_{A}]}))$ outperform the rest when $n_{X}>40$.

Other unexpected behaviours observed in the {\it WikiCS} dataset are depicted in figures~\ref{fig:w_GCN_X_c_A_c} and~\ref{fig:w_GCN_X_c_A_z}.

% X_z
\paragraph{Blanked out node feature matrix ($GCN:X_{z}$)}
\paragraph{ogbn-arxiv dataset}
Under this condition of distortion, we observe the following: 
(1) The layers can hardly perform any denoising,
(2) $\mathcal{P}(\psi_{0}^{GCN}(X_{z}^{[n_{X}]},A^{[n_{A}]}))$ shows a linear downward decrease in performance,
(3) $\mathcal{P}(\psi_{3}^{GCN}(X_{z}^{[n_{X}]},A^{[n_{A}]}))$ usually performs poorly although in some instances, it does some denoising, 
(4) At high values of $n_{A}$, $\mathcal{P}(\psi_{1}^{GCN}(X_{z}^{[n_{X}]},A^{[n_{A}]}))$ performs poorly with some denoising ability.

% (X_z,A_z) > (X_z,A_c)
When the adjacency matrix is blanked out rather than corrupted, we observe that the decrease in accuracy is much more gradual in the former as compared to the latter.

\paragraph{WikiCS dataset}
Here we see the following:
(1) $\mathcal{P}(\psi_{0}^{GCN}(X_{z}^{[n_{X}]},A^{[n_{A}]}))$ is able to perform some analysis but exhibits a bomb-like trajectory (similar to figure~\ref{fig:n2v_X_z_A_c_0}) at high values of $n_{X}$,
(2) $\mathcal{P}(\psi_{1}^{GCN}(X_{z}^{[n_{X}]},A^{[n_{A}]}))$ initially performs poorly but picks up as $n_{A}$ increases,
(3) As $n_{X}$ increases, the behaviour of the layers start resembling that of {\it ogbn-arxiv} dataset.

\begin{figure}[htpb]
    \centering
    \begin{minipage}{0.5\textwidth}
        \centering
        \includegraphics[scale=0.25]{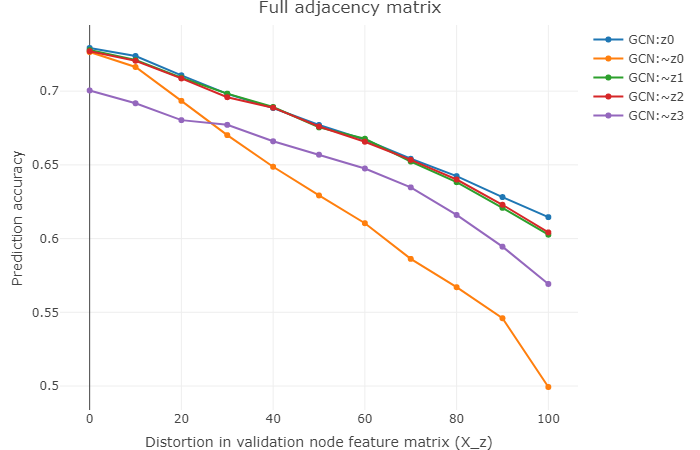}
        \subcaption{{\it ogbn-arxiv} dataset \label{fig:sub:o_GCN_X_z_A_c_0}}
    \end{minipage}%
    \begin{minipage}{0.5\textwidth}
        \centering
        \includegraphics[scale=0.25]{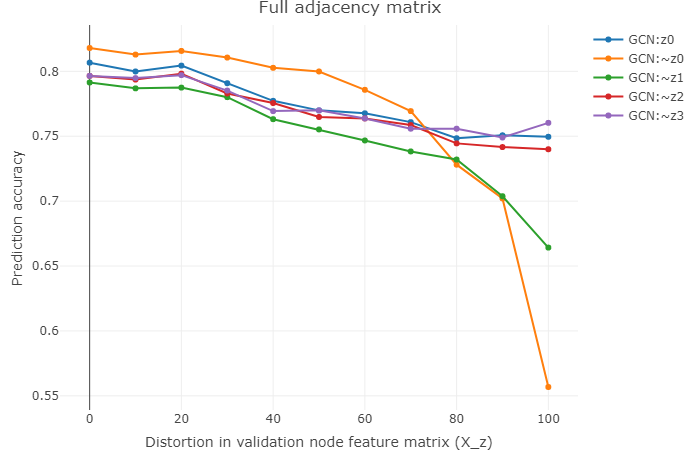}
        \subcaption{{\it WikiCS} dataset\label{fig:sub:w_GCN_X_z_A_c_0}}
    \end{minipage}
    \caption{Representative behaviour of $\mathcal{P}(\psi_{i}^{GCN}(X_{z}^{[n_{X}]},A_{c}^{[0]}))$ and $\mathcal{P}(\phi^{GCN}(X_{z}^{[n_{X}]},A_{c}^{[0]})), i \in \{0,1,2,3\}, n_{X} \in \{0,10,\dots,100\}$.}
    \label{fig:GCN_X_z_A_c_0}
\end{figure}

Figure~\ref{fig:GCN_X_z_A_c_0} is representative of the behaviour which is exhibited when the node feature matrix is blanked out in GCN-based denoising pipeline.

\subsubsection{GraphSAGE-based denoising pipeline}\label{sec:SAGE}
\paragraph{Corrupted node feature matrix ($SAGE:X_{c}$)}
In the case of the {\it ogbn-arxiv} dataset, we see that: $\mathcal{P}(\psi_{1}^{SAGE}(X_{c}^{[n_{X}]},A^{[n_{A}]}))$ outperforms the rest, which is closely followed by $\mathcal{P}(\psi_{0}^{SAGE}(X_{c}^{[n_{X}]},A^{[n_{A}]}))$.
For {\it WikiCS} dataset, we observe that $\mathcal{P}(\psi_{0}^{SAGE}(X_{c}^{[n_{X}]},A^{[n_{A}]}))$ performs a lot better than the rest of the layers, while $\mathcal{P}(\psi_{1}^{SAGE}(X_{c}^{[n_{X}]},A^{[n_{A}]}))$ attempts to perform some denoising.

\begin{figure}[htpb]
    \centering
    \begin{minipage}{0.5\textwidth}
        \centering
        \includegraphics[scale=0.25]{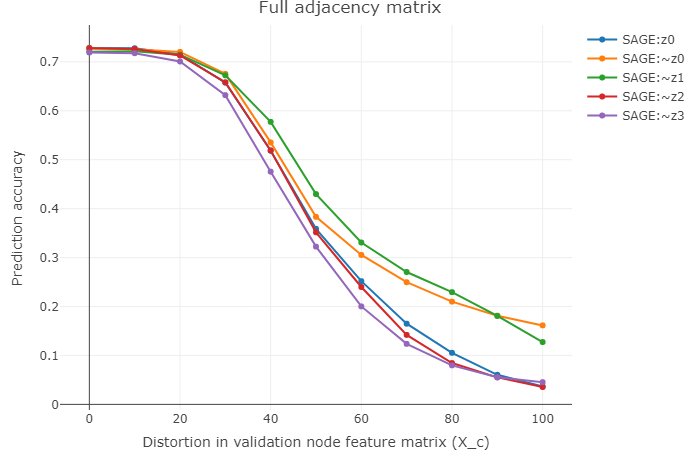}
        \subcaption{{\it ogbn-arxiv} dataset \label{fig:sub:o_SAGE_X_c_A_c_0}}
    \end{minipage}%
    \begin{minipage}{0.5\textwidth}
        \centering
        \includegraphics[scale=0.25]{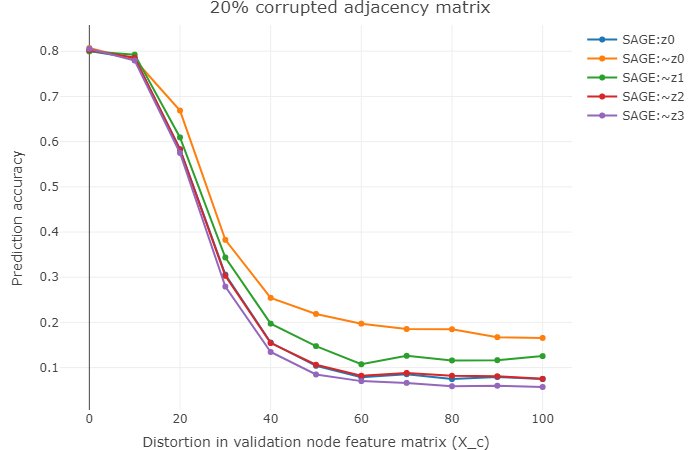}
        \subcaption{{\it WikiCS} dataset \label{fig:sub:w_SAGE_X_c_A_c_20}}
    \end{minipage}
    \caption{Representative behaviour of $\mathcal{P}(\psi_{i}^{SAGE}(X_{c}^{[n_{X}]},A^{[n_{A}]}))$ and $\mathcal{P}(\phi^{SAGE}(X_{c}^{[n_{X}]},A^{[n_{A}]})), i \in \{0,1,2,3\}, n_{X} \in \{0,10,\dots,100\}, n_{A} \in \{0,10,\dots,100\}$.}
    \label{fig:SAGE_X_c_A_c}
\end{figure}

\paragraph{Blanked out node feature matrix ($SAGE:X_{z}$)}
We observe that: \\ $\mathcal{P}(\psi_{0}^{SAGE}(X_{z}^{[n_{X}]},A^{[n_{A}]}))$ performs the best.
This is followed by \\ $\mathcal{P}(\psi_{1}^{SAGE}(X_{z}^{[n_{X}]},A^{[n_{A}]}))$.
In the case of {\it ogbn-arxiv} dataset, we observe that $\mathcal{P}(\psi_{3}^{SAGE}(X_{z}^{[n_{X}]},A^{[n_{A}]}))$ consistently exhibits poor performance, while in {\it WikiCS} dataset, we see that $\mathcal{P}(\psi_{2}^{SAGE}(X_{z}^{[n_{X}]},A^{[n_{A}]}))$, $\mathcal{P}(\psi_{3}^{SAGE}(X_{z}^{[n_{X}]},A^{[n_{A}]}))$ and $\mathcal{P}(\phi^{SAGE}(X_{z}^{[n_{X}]},A^{[n_{A}]}))$ form a curve that becomes increasingly convex as value of $n_{A}$ increases.

\begin{figure}[htpb]
    \centering
    \begin{minipage}{0.5\textwidth}
        \centering
        \includegraphics[scale=0.25]{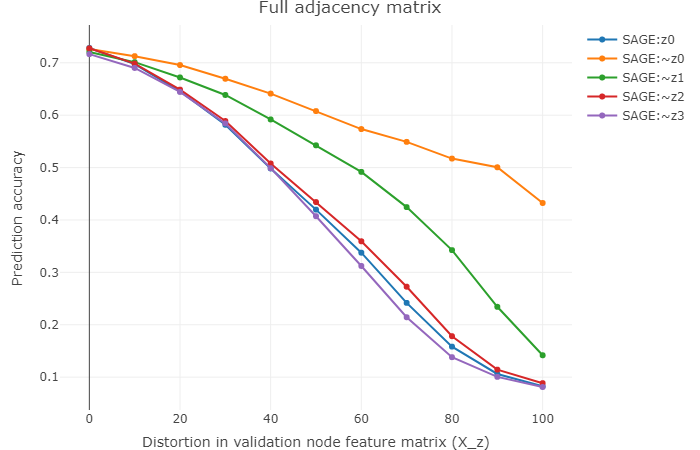}
        \subcaption{{\it ogbn-arxiv} dataset \label{fig:sub:o_SAGE_X_z_A_z_0}}
    \end{minipage}%
    \begin{minipage}{0.5\textwidth}
        \centering
        \includegraphics[scale=0.25]{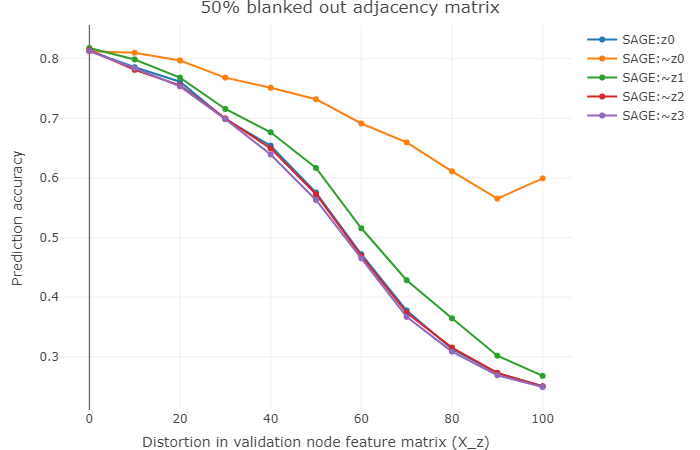}
        \subcaption{{\it WikiCS} dataset \label{fig:sub:w_SAGE_X_z_A_z_50}}
    \end{minipage}
    \caption{Representative behaviour of $\mathcal{P}(\psi_{i}^{SAGE}(X_{z}^{[n_{X}]},A^{[n_{A}]}))$ and $\mathcal{P}(\phi^{SAGE}(X_{z}^{[n_{X}]},A^{[n_{A}]})), i \in \{0,1,2,3\}, n_{X} \in \{0,10,\dots,100\}, n_{A} \in \{0,10,\dots,100\}$.}
    \label{fig:SAGE_X_z_A_z}
\end{figure}
%%%%%%%%%%%%%%%%%%%%%%%%%%%%%%%%%%%%%%%%%%%%%%%%%%%%%%%%%%%%%%
% Discussions
%%%%%%%%%%%%%%%%%%%%%%%%%%%%%%%%%%%%%%%%%%%%%%%%%%%%%%%%%%%%%%
\section{Discussion}\label{sec:discuss}

In this section we will attempt to explain the behavior of the layers in each of the denoising pipelines.
Recall from section~\ref{sec:DNN_MI} that decrease in mutual information indicates an increase in prediction accuracy which is in turn driven by the phenomenon of progressive geometric clustering of the samples belonging to the same class.

\subsection{Issues regarding estimation of mutual information}\label{sec:MI_bias}
% bias can't be computed
\cite{goldfeld2018entropy} argues that while the variance of the estimate of differential entropy term can be empirically evaluated using bootstrapping, there is no empirical test for bias. 
% #(samples) exponential in #(dimensions) for small bias
Theorem 4 therein states that for negligible bias, the number of samples must be atleast $2^{0.99d}$ where $d$ denotes the dimensionality of $X$. 
% convergence can be deceptive
Furthermore, even if multiple estimates of the differential entropy term consistently produce similar values, this does not necessarily suggest that these values are closer to the true entropy value.
% reducing d to log2(N)/0.99 is not going to help
However, reduction in dimensionality affects both our denoising task and the ability of RBM to reconstruct the embeddings, thereby directly affecting the prediction accuracy.
% remind readers of ogbn-arxiv
Recall that in the {\it ogbn-arxiv} dataset, we have $128$ dimensional node feature vectors and $256$ dimensional hidden layer representations (except for the final layer) (refer section~\ref{sec:ogbn-arxiv}).
% impossible to get 2^{0.99*256} and compute
Given this information, it is both impossible to obtain the desired number of samples and compute the required estimation using these many samples.
% high bias problem
Computation with the available $169343$ nodes will result in an estimation of $I(X;T_{l})$ for each layer $l$ that will most likely have high bias.
% error will be very significant to disregard
This is not reflective of the right behavior and may not be consistent with the observed clustering of samples.
% theoretically can't reduce number of samples to a number practical for graph datasets
Additionally, the requirement of maintaining a low-$\beta$ regime to observe the relation between geometric clustering and reduction in mutual information~\ref{sec:intro} prevents a large reduction in the desired number of samples.

% Why choose WikiCS?
Due to these reasons, the estimation of $I(X:T_{l})$ for {\it ogbn-arxiv} dataset does not seem practical.
Additionally, tracking geometric clustering in $256$ dimensional space is cumbersome.
To counter some of these issues, we chose the {\it WikiCS} dataset which has $300$ dimensional node feature vectors with $35$ dimensional hidden layer representations (except for the final layer) (refer section~\ref{sec:wikics}).
Although this dimensionality still requires a very large number of samples for estimation of $I(X:T_{l})$ for each layer $l$, and also suffers from high bias, the {\it WikiCS} dataset presents two advantages: 
(1) two dimensional t-SNE embeddings from these $35$ dimensional representations will be easier to view, 
(2) validation set consists of just $1769$ nodes in the validation set and $5847$ nodes in the test set, hence subtle clustering in the representation space is much more evident.

Therefore, we will rely on this progressive geometric clustering exhibited by {\it WikiCS} dataset rather than the very likely erroneous estimate of the desired mutual information term to explain our findings.
Regrettably, this indicates that we will only be able to explain the observations exhibited by {\it WikiCS} dataset and those that match in {\it ogbn-arxiv} dataset.

\subsection{Explanation of behavior}\label{sec:explain}
Here we attempt to explain the behavior of the layers of the various denoising pipelines for the {\it WikiCS} dataset.

We encourage the reader to load the web application for the corresponding denoising pipeline (refer section~\ref{sec:cluster}) alongside this section since we will be referring to those plots in our explanation. 

From the {\it desired representations}, t-SNE embeddings of each of the layers can be described as:
\begin{enumerate}
    \item $z_{0}$: The representation in the input space depict the inherent structure of the dataset.
    \item $z_{1}$ ({\it re-positioning phase}): The first hidden layer representation shows a re-positioned view of the samples where those belonging to the same class are positioned closer to one another.
    \item $z_{2}$ ({\it clustering phase}): In the second hidden layer representation, the samples belonging to the same class form clusters.
    \item $z_{3}$ ({\it tightening phase}): In the final layer representation, these clusters tighten thereby making samples from the same class indistinguishable with visible decision boundaries. 
    This makes the clusters maximally linearly separable, thus resulting in good prediction accuracy.
\end{enumerate}

Recall from section~\ref{sec:MI_bias} that we can only look at the observations exhibited by {\it WikiCS} dataset and those that are in common with {\it ogbn-arxiv} dataset.

% What does the RBM do?
The task of {\it RBM}-$z_{i}$ is to use the representations of $NN:z_{0}\ z_{i}$ to generate a reconstruction $NN:\sim z_{i}\ z_{i}$ that is close in structure to the desired representation $z_{i}$.
As distortion in the data increases, the representations of $NN:z_{i}\ z_{i}$ keeps getting farther away from $z_{i}$.
Farther the reconstructions are from the desired representation, higher is the error in reconstruction.

% RBM and DNN go hand in hand
Additionally, it not only matters how close the reconstructions are to the desired representations.
It also depends on how many hidden layers are remaining in the DNN from the layer where reconstructions are fed back.
Reconstruction errors in the initial hidden layers can be alleviated sequentially through the hidden layers that follow.
Therefore, denoising ability of reconstructions of layers decreases as we move down the network.

In this section we will look at the typical behavior of the denoising pipelines and consider some atypical cases of interest in the appendix.

The behavior of the GCN-based denoising pipeline is not very consistent to be analyzed under this framework.
Estimation of $I(X:T_{l})$ for each layer $l$ is required for analyzing the behavior in the pipeline.

% NN:z_0 representations
\subsubsection{$NN:z_{0}$ representations}
% NN:z_0 z_0
With increasing distortion in the data, the representations of $NN:z_{0}\ z_{0}$ gradually lose their structure which can be categorized as: 
(1) $X_{c}$: as $n_{X}$ increases, the samples begin to move apart into a circular assortment of points from random classes, 
(2) $X_{z}$: with increasing $n_{X}$, we observe that the samples form groups of datapoints from arbitrary classes, 
(3) $A$: a rise in $n_{A}$ causes the samples from the periphery of the arrangement to disperse. $A_{z}$ causes more dispersion than $A_{c}$.

% NN:z_0 z_i -> NN:z_0 z_(i+1)
Each step of distortion in $NN:z_{0}\ z_{i}$ causes $NN:z_{0}\ z_{(i+1)}$, $i \in \{0,1,2\}$ to move towards one of the following:
(1) $X_{c}$: a circular assortment of random samples that is slightly more clustered,
(2) $X_{z}$: groups of samples of random classes that first undergo unraveling followed by some clustering,
(3) $A$: a dispersed representation which is more clustered.
This moves $NN:z_{0}\ z_{3}$ towards a more clustered version of a random representation of samples in $NN:z_{0}\ z_{0}$, leading to a decline in $\mathcal{P}(\phi^{NN}(X^{[n_{X}]}[,A^{[n_{A}]}]))$.

\subsubsection{$NN:\sim z_{0}$ representations}
% NN:~z_0 z_0
We observe that {\it RBM}-$z_{0}$ is able to generate a reconstruction that usually resembles $z_{0}$ more than $NN:\sim z_{0}\ z_{0}$ does.
This means the following:
(1) $X_{c}$: the samples from the random classes in the circular arrangement are drawn towards the inherent structure in $z_{0}$,
(2) $X_{z}$: the groups of random samples are unwound and moved towards $z_{0}$,
(3) $A$: the samples that have dispersed from the periphery are brought closer.
% NN:~z_0 z_1
This helps $NN_{1}$ generate a better re-positioned representation $NN:\sim z_{0}\ z_{1}$ than $NN:z_{0}\ z_{1}$.
% NN:~z_0 z_2
This advantage is passed on with better clustering by $NN_{2}$ 
% NN:~z_0 z_3
and a tighter representation $NN:\sim z_{0}\ z_{3}$ by $NN_{3}$.
This is the reason that $\mathcal{P}(\psi_{0}^{NN}(X^{[n_{X}]}[,A^{[n_{A}]}]))$ is improved over $\mathcal{P}(\phi^{NN}(X^{[n_{X}]}[,A^{[n_{A}]}]))$.

\subsubsection{$NN:\sim z_{1}$ representations}
% NN:~z_1 z_1
The samples from RBM-$z_{1}$ after being fed $NN:\sim z_{1}\ z_{1}$ is drawn closer to $z_{1}$ but is not as well re-positioned as $NN:\sim z_{0}\ z_{1}$.
% NN:~z_1 z_2 and NN:~z_1 z_3
In general after processing by $NN_{2}$ and $NN_{3}$, we observe that $NN:\sim z_{1}\ z_{3}$ has greatly improved when compared with $NN:\sim z_{0}\ z_{3}$ but not as well clustered as $NN:\sim z_{0}\ z_{3}$.
Therefore, even if $\mathcal{P}(\psi_{1}^{NN}(X^{[n_{X}]}[,A^{[n_{A}]}]))$ is better than $\mathcal{P}(\phi^{NN}(X^{[n_{X}]}[,A^{[n_{A}]}]))$ but still lower than $\mathcal{P}(\psi_{0}^{NN}(X^{[n_{X}]}[,A^{[n_{A}]}]))$.

\paragraph{MLP:$X_{c}$}
We observe that the representations of $MLP:\sim z_{1}$ become comparable with those of $MLP:\sim z_{0}$ when $n_{X} > 40$.
This is why $\mathcal{P}(\psi_{1}^{MLP}(X_{c}^{[n_{X}]},\cdot))$ becomes about the same as $\mathcal{P}(\psi_{0}^{MLP}(X_{c}^{[n_{X}]},\cdot))$.

\paragraph{$X_{z}$}
In this case, RBM-$z_{1}$ has to deal with unraveling the group of samples in $NN:z_{0}\ z_{1}$ before re-positioning the points.
The errors in this re-positioning affect the clustering ability of $NN_{2}$.
Hence the representations of $NN_{3}$, although better than $NN:z_{0}\ z_{3}$, are not as well-clustered as $NN:\sim z_{0}\ z_{3}$.

\subsubsection{$NN:\sim z_{2}$ representations}\label{sec:NN_sim_z2}
% NN:~z_2 z_2
The reconstructions of {\it RBM}-$z_{2}$ hardly show any noticeable difference from $NN:\sim z_{0}\ z_{2}$.
The final hidden layer, acting simply as a linear classifier, is unable to improve this representation.
Therefore, $\mathcal{P}(\psi_{2}^{NN}(X^{[n_{X}]}[,A^{[n_{A}]}]))$ is very similar to $\mathcal{P}(\phi^{NN}(X^{[n_{X}]}[,A^{[n_{A}]}]))$.
The deterioration of $\mathcal{P}(\psi_{2}^{NN}(X^{[n_{X}]}[,A^{[n_{A}]}]))$ in $MLP:X_{c}$ is not substantially reflected in the representational changes of $MLP:\sim z_{2}\ z_{2}$. 
Estimation of $I(X:T_{2})$ is necessary to shed light on such observations.

\subsubsection{$NN:\sim z_{3}$ representations}
% NN:~z_3 z_3
The behavior of RBM-$z_{3}$ and $\mathcal{P}(\psi_{3}^{NN}(X^{[n_{X}]}[,A^{[n_{A}]}]))$ is same as that of section~\ref{sec:NN_sim_z2}.
%%%%%%%%%%%%%%%%%%%%%%%%%%%%%%%%%%%%%%%%%%%%%%%%%%%%%%%%%%%%%%
% Related work
%%%%%%%%%%%%%%%%%%%%%%%%%%%%%%%%%%%%%%%%%%%%%%%%%%%%%%%%%%%%%%
\section{Related work} %(\textit{EF: This section is too short and shallow}) (AM: I have edited this section. Please have a look.)}
\label{sec:related}
\cite{chen2020graph} proposed a GNN framework to denoise single and multiple noisy graph signals.
They unroll an iterative denoising algorithm by mapping each iteration into a single network layer where the feed-forward process is equivalent to iteratively denoising graph signals.
The graph unrolling networks are trained through unsupervised learning, where the input noisy graph signals are used to supervise the networks.
This way the networks can adaptively capture appropriate priors from input noisy graph signals, instead of manually choosing signal priors which are usually too complicated to be explicitly and precisely described in mathematical terms or may lead to complicated and computationally intensive algorithms.
The convolution operation used in the graph unrolling network is permutation equivariant and can flexibly adjust the edge weights to various graph signals.
\cite{chen2014signal} focuses on denosing graph signals from noisy measurements. 
They consider graph signal denoising as an optimization problem based on regularization of graph total variation of noisy signals. 
An exact closed-form solution expressed by an inverse graph filter is proposed.
Since this solution requires a matrix inversion which is $O(N^{3})$ where $N$ is the number of nodes, an approximate iterative solution expressed by a standard graph filter is derived which is $O(LKN)$ where $L$ is the number of filter taps and $K$ is the number of non-zero elements.
\cite{Chen_2015} looks at graph signal recovery as an optimization problem, for which we provide a general solution through the alternating direction methods of multipliers.

\cite{wang2019learning} proposed {\it Graph Denoising Policy Network} (GDPNet) to learn robust representations from noisy graph data through reinforcement learning.
They argue that since GNNs rely on aggregation of neighbourhood information, these models are vulnerable to noises present in the input graph.
GDPNet has two phases: {\it Signal neighborhood selection phase} and {\it Representation learning phase}.
In the signal neighborhood selection phase, for each target node, the noisy neighbors are removed by formulating this removal process as a Markov decision process (MDP) solved through {\it policy gradients}.
The corresponding policy which determines this process is decided with task-specific rewards received from the representation learning phase. 
The remaining nodes are called the {\it signal neighbors} of the target node. 
The information from these signal neighbors are aggregated to learn node representations for downstream tasks.
The performance of the resulting node representations provides task-specific rewards for the signal neighborhood selection phase.
These two phases are jointly trained to obtain the optimal neighbor set for target nodes with maximum cumulative task-specific rewards and robust representations for the nodes.
    
\cite{fu2020understanding} provides a theoretical framework to understand GNNs from a graph signal denoising perspective.
This framework shows that GNNs are implicitly solving graph signal denoising problems: spectral graph convolutions work as denoising node features, while graph attentions work as denoising edge weights.
Their results lead to models which work effectively for graphs with noisy node features and/or noisy edges by working through a tradeoff between node feature denoising and smoothing.
Instead of extracting high-level features, spectral graph convolution operators are simply denoising and smoothing the input node features.
\cite{nt2019revisiting} find that the feature vectors of benchmark datasets are already quite informative for the classification task, and the graph structure only provides a means to denoise the data. 
They develop a theoretical framework based on graph signal denoising perspective for analyzing graph neural networks.
The results indicate that graph neural networks only perform low-pass filtering on feature vectors and do not have the non-linear manifold learning property.

Unlike these works, our paper does not deal with denoising the noisy data directly but capitalizes on models that have been trained on clean training data.
%%%%%%%%%%%%%%%%%%%%%%%%%%%%%%%%%%%%%%%%%%%%%%%%%%%%%%%%%%%%%%
% Conclusions
%%%%%%%%%%%%%%%%%%%%%%%%%%%%%%%%%%%%%%%%%%%%%%%%%%%%%%%%%%%%%%
\section{Conclusions and future work}\label{sec:conclude}
In this paper, we propose a denoising framework that exploits the associative memory property of an RBM and the hidden layer representations of a DNN.
We have seen that a DNN trained on clean training dataset can be used to denoise a variety of distortions in unseen data.
We believe that this skips the need to alter a trained DNN that has been deployed into production in an expert system.
By training an RBM on the chosen hidden layer representations, the DNN is largely robust to noise occuring at test time.

We have shown that training an RBM on the representations of the first hidden layer performs the most denoising.
When this is not the case, an RBM trained on the node feature matrix is able to denoise the unseen data considerably.
The reader is encouraged to use the {\it playground} functionality available at \url{https://ankithmo.shinyapps.io/denoiseRBM} to compare the performance of the various models at specific values of noise.

We offer a peek into the workings of the DNNs used in our pipeline by observing the t-SNE embeddings of the underlying hidden layer representations.
By establishing that a rise in prediction accuracy is driven by a tightening in geometric clustering of these hidden layer representations, we attempt to explain why such behaviors are depicted by our denoising pipelines.

Although the estimation of $I(X:T_{l})$ is a more suitable measure for tracking the increase in accuracy, we have seen that its estimation from the available samples is computationally very difficult.
Even if such an estimate was computed from the available samples, this measure would have very high bias.
Understanding why such differences exist between the reconstructions of an RBM requires rigorous mathematical analysis which we defer to our future work.

% when training data not available
In general GB-RBMs are not robust to noise as it assumes a diagonal Gaussian as its conditional distribution over the visible nodes. 
This means that the log probability assigned to a noisy outlier would be very low and classification accuracy tends to be poor for noisy, out-of-sample test cases. 
We intend to investigate RBMs that have visible units with other suitable distributions for our denoising task.
Robust restricted Boltzmann machine (RoBM) proposed by~\cite{tang2012robust} have been shown to be robust to corruptions in the training set and are capable of accurately dealing with occlusions and noise by using multiplicative gating to induce a scale mixture of Gaussians over pixels.
RoBM have been successfully used in image denoising and inpainting. 
In cases where clean training set is not available, such RoBMs can be used for on-the-fly denoising tasks.

% future work
We hope to extend these denoising tasks to other datasets, other ML tasks, and other features related to nodes and edges, such as distortions in the node position matrix, edge feature matrix, edge position matrix, etc.
%%%%%%%%%%%%%%%%%%%%%%%%%%%%%%%%%%%%%%%%%%%%%%%%%%%%%%%%%%%%%%
% Bibliography
%%%%%%%%%%%%%%%%%%%%%%%%%%%%%%%%%%%%%%%%%%%%%%%%%%%%%%%%%%%%%%
\bibliographystyle{unsrt}
\bibliography{references}
%%%%%%%%%%%%%%%%%%%%%%%%%%%%%%%%%%%%%%%%%%%%%%%%%%%%%%%%%%%%%%
% Appendix
%%%%%%%%%%%%%%%%%%%%%%%%%%%%%%%%%%%%%%%%%%%%%%%%%%%%%%%%%%%%%%
\newpage
\appendix
\section{Appendix}
This Appendix presents additional plots that exhibit notable atypical behaviour exhibited by the node2vec-, GCN- and GraphSAGE-based denoising pipelines.

\subsection{node2vec-based denoising pipelines}
% representative case figure
\begin{figure}[h]
    \centering
    \begin{minipage}{0.5\textwidth}
        \centering
        \includegraphics[scale=0.25]{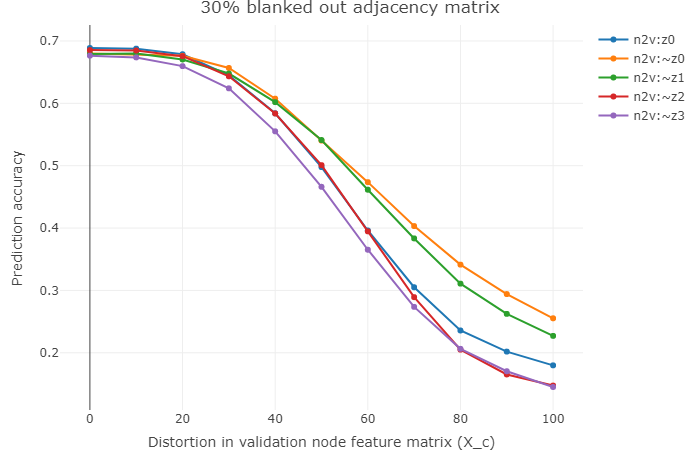}
        \subcaption{{\it ogbn-arxiv} dataset \label{fig:sub:o_n2v_X_c_A_z}}
    \end{minipage}%
    \begin{minipage}{0.5\textwidth}
        \centering
        \includegraphics[scale=0.25]{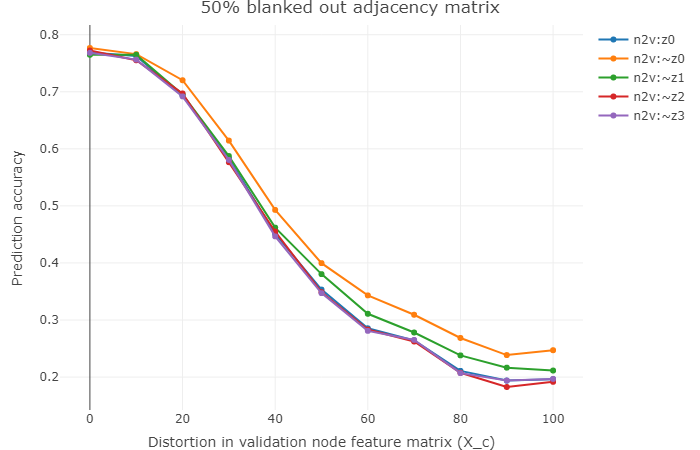}
        \subcaption{{\it WikiCS} dataset\label{fig:sub:w_n2v_X_c_A_z}}
    \end{minipage}
    \caption{Representative behaviour of $\mathcal{P}(\psi_{i}^{n2v}(X_{c}^{[n_{X}]},A_{z}^{[n_{A}]}))$ and $\mathcal{P}(\phi^{n2v}(X_{c}^{[n_{X}]},A_{z}^{[n_{A}]})), i \in \{0,1,2,3\}, n_{X} \in \{0,10,\dots,100\}, n_{A} \in \{0,10,\dots,100\}$.} 
    \label{fig:n2v_X_c_A_z}
\end{figure}

\begin{figure}[h]
    \centering
    \includegraphics[scale=0.34]{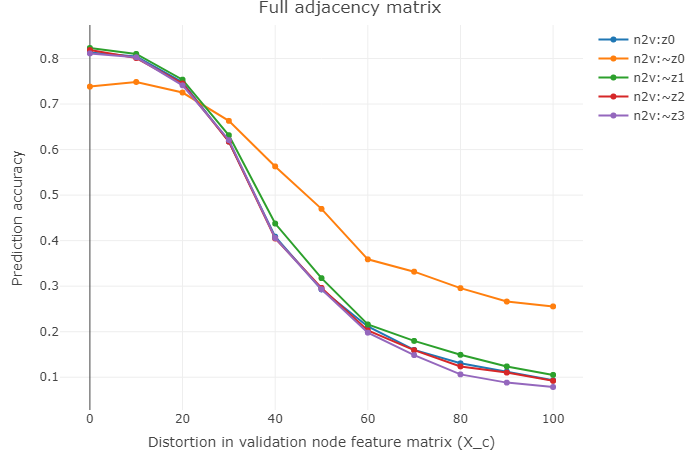}
    \caption{Atypical behaviour of $\mathcal{P}(\psi_{i}^{n2v}(X_{c}^{[n_{X}]},A_{c}^{[0]}))$ and $\mathcal{P}(\phi^{n2v}(X_{c}^{[n_{X}]},A_{c}^{[0]})), i \in \{0,1,2,3\}, n_{X} \in \{0,10,\dots,100\}$ for WikiCS dataset.}
    \label{fig:n2v_X_c_A_c_0}
\end{figure}

\begin{figure}[h]
    \centering
    \includegraphics[scale=0.34]{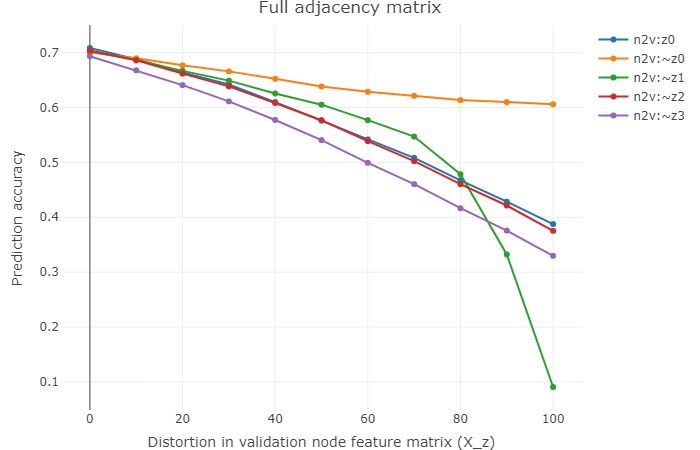}
    \caption{Atypical behaviour of $\mathcal{P}(\psi_{i}^{n2v}(X_{z}^{[n_{X}]},A_{c}^{[0]}))$ and $\mathcal{P}(\phi^{n2v}(X_{z}^{[n_{X}]},A_{c}^{[0]})), i \in \{0,1,2,3\}, n_{X} \in \{0,10,\dots,100\}$ for {\it ogbn-arxiv} dataset.}
    \label{fig:n2v_X_z_A_c_0}
\end{figure}

Figures~\ref{fig:n2v_X_c_A_c_0} and~\ref{fig:n2v_X_z_A_c_0} show atypical behaviour of node2vec-based denoising pipelines.

~

\subsection{GCN-based denoising pipelines}
\begin{figure}[htpb]
    \centering
    \begin{minipage}{0.5\textwidth}
        \centering
        \includegraphics[scale=0.25]{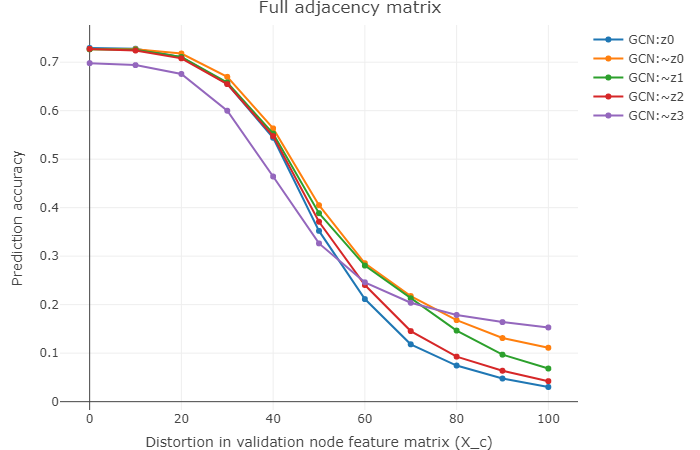}
        \subcaption{{\it ogbn-arxiv dataset} \label{fig:sub:o_GCN_X_c_A_c_0}}
    \end{minipage}%
    \begin{minipage}{0.5\textwidth}
        \centering
        \includegraphics[scale=0.25]{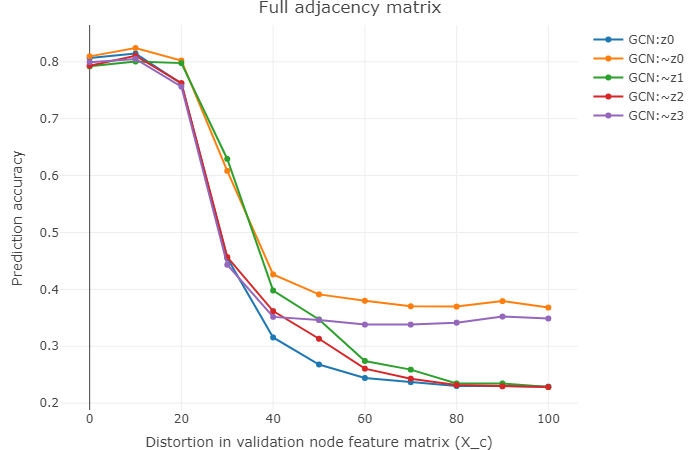}
        \subcaption{{WikiCS dataset}\label{fig:sub:w_GCN_X_c_A_c_0}}
    \end{minipage}
    \caption{Atypical behaviour of $\mathcal{P}(\psi_{i}^{GCN}(X_{c}^{[n_{X}]},A_{c}^{[0]}))$ and $\mathcal{P}(\phi^{GCN}(X_{c}^{[n_{X}]},A_{c}^{[0]})), i \in \{0,1,2,3\}, n_{X} \in \{0,10,\dots,100\}$.}
    \label{fig:GCN_X_c_A_c_0}
\end{figure}

\begin{figure}[htpb]
    \centering
    \begin{minipage}{0.5\textwidth}
        \centering
        \includegraphics[scale=0.25]{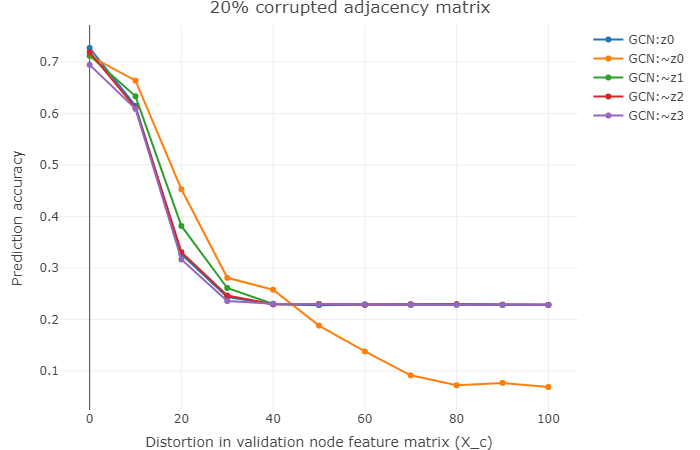}
        \subcaption{\label{fig:sub:w_GCN_X_c_A_c_20}}
    \end{minipage}%
    \begin{minipage}{0.5\textwidth}
        \centering
        \includegraphics[scale=0.25]{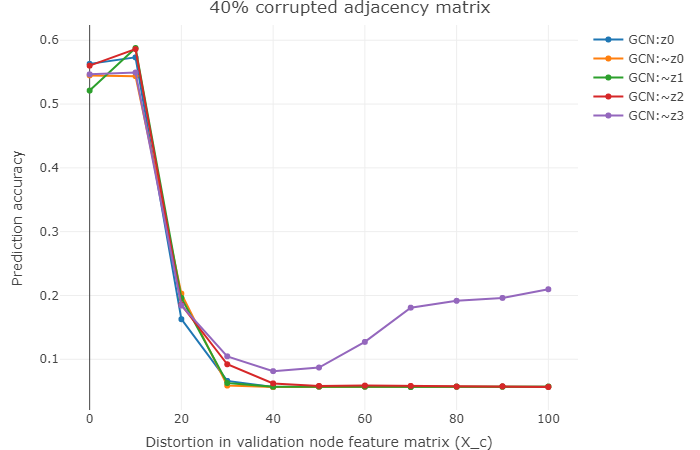}
        \subcaption{\label{fig:sub:w_GCN_X_c_A_c_40}}
    \end{minipage}
    \begin{minipage}{0.5\textwidth}
        \centering
        \includegraphics[scale=0.25]{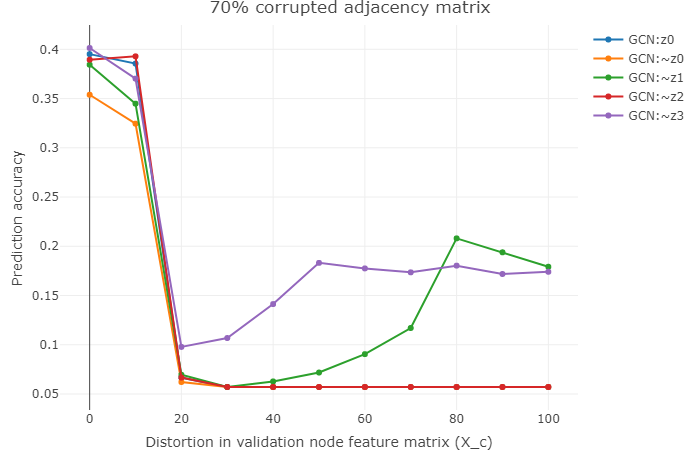}
        \subcaption{\label{fig:sub:w_GCN_X_c_A_c_70}}
    \end{minipage}%
    \begin{minipage}{0.5\textwidth}
        \centering
        \includegraphics[scale=0.25]{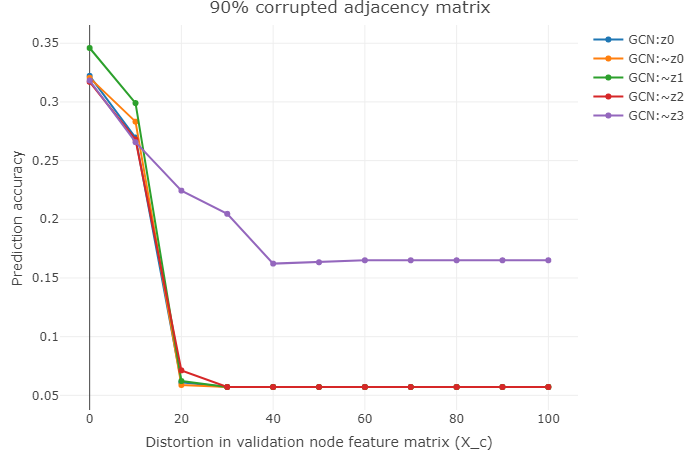}
        \subcaption{\label{fig:sub:w_GCN_X_c_A_c_90}}
    \end{minipage}
    \caption{Atypical behaviour of $\mathcal{P}(\psi_{i}^{GCN}(X_{c}^{[n_{X}]},A_{c}^{[n_{A}]}))$ and $\mathcal{P}(\phi^{GCN}(X_{c}^{[n_{X}]},A_{c}^{[n_{A}]})), i \in \{0,1,2,3\}, n_{X} \in \{0,10,\dots,100\}, n_{A} \in \{20,40,70,90\}$ for {\it WikiCS} dataset.}
    \label{fig:w_GCN_X_c_A_c}
\end{figure}

\begin{figure}[htpb]
    \centering
    \begin{minipage}{0.5\textwidth}
        \centering
        \includegraphics[scale=0.25]{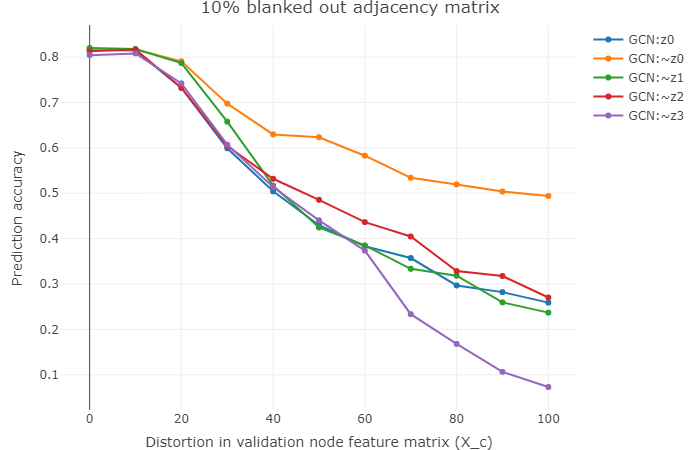}
        \subcaption{\label{fig:sub:w_GCN_X_c_A_z_10}}
    \end{minipage}%
    \begin{minipage}{0.5\textwidth}
        \centering
        \includegraphics[scale=0.25]{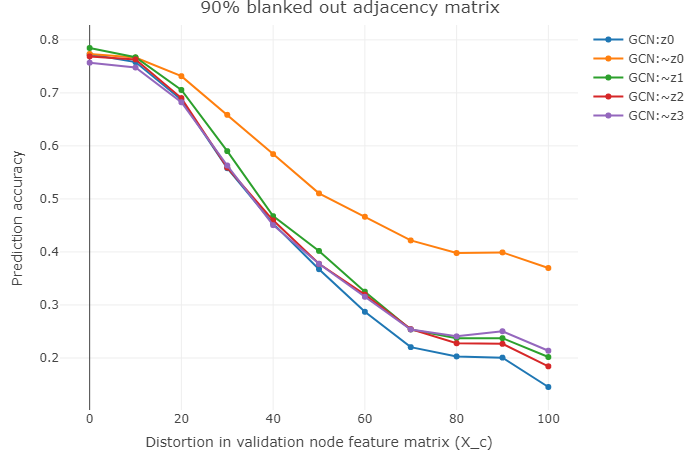}
        \subcaption{\label{fig:sub:w_GCN_X_c_A_z_90}}
    \end{minipage}
    \caption{Atypical behaviour of $\mathcal{P}(\psi_{i}^{GCN}(X_{c}^{[n_{X}]},A_{z}^{[n_{A}]}))$ and $\mathcal{P}(\phi^{GCN}(X_{c}^{[n_{X}]},A_{z}^{[n_{A}]})), i \in \{0,1,2,3\}, n_{X} \in \{0,10,\dots,100\}, n_{A} \in \{10,90\}$ for {\it WikiCS} dataset.}
    \label{fig:w_GCN_X_c_A_z}
\end{figure}

\begin{figure}[htpb]
    \centering
    \begin{minipage}{0.33\textwidth}
        \centering
        \includegraphics[scale=0.175]{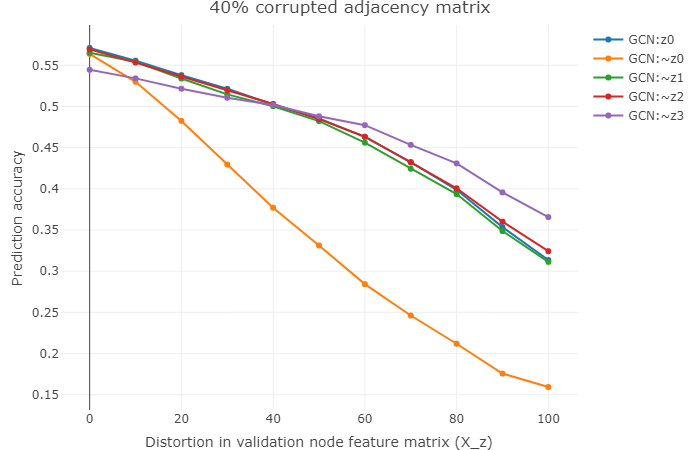}
        \subcaption{\label{fig:sub:o_GCN_X_z_A_c_40}}
    \end{minipage}%
    \begin{minipage}{0.33\textwidth}
        \centering
        \includegraphics[scale=0.175]{figs/ogbn_arxiv/GCN/o_GCN_X_z_A_c_40.png}
        \subcaption{\label{fig:sub:o_GCN_X_z_A_c_60}}
    \end{minipage}%
    \begin{minipage}{0.33\textwidth}
        \centering
        \includegraphics[scale=0.175]{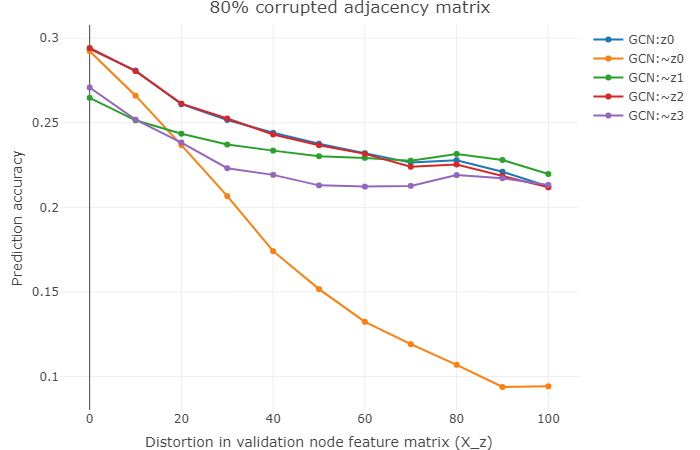}
        \subcaption{\label{fig:sub:o_GCN_X_z_A_c_80}}
    \end{minipage}
    \caption{Atypical behaviour of $\mathcal{P}(\psi_{i}^{GCN}(X_{z}^{[n_{X}]},A_{c}^{[n_{A}]}))$ and $\mathcal{P}(\phi^{GCN}(X_{z}^{[n_{X}]},A_{c}^{[n_{A}]})), i \in \{0,1,2,3\}, n_{X} \in \{0,10,\dots,100\}, n_{A} \in \{40,60,80\}$ for {\it ogbn-arxiv} dataset.}
    \label{fig:o_GCN_X_z_A_c}
\end{figure}

\begin{figure}[htpb]
    \centering
    \begin{minipage}{0.5\textwidth}
        \centering
        \includegraphics[scale=0.25]{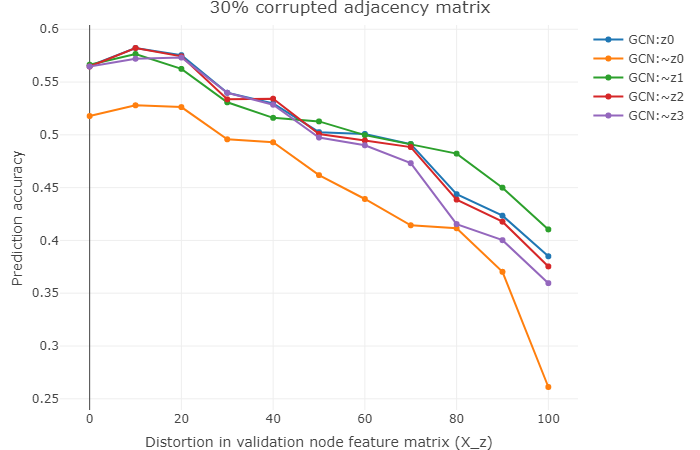}
        \subcaption{\label{fig:sub:w_GCN_X_z_A_c_30}}
    \end{minipage}%
    \begin{minipage}{0.5\textwidth}
        \centering
        \includegraphics[scale=0.25]{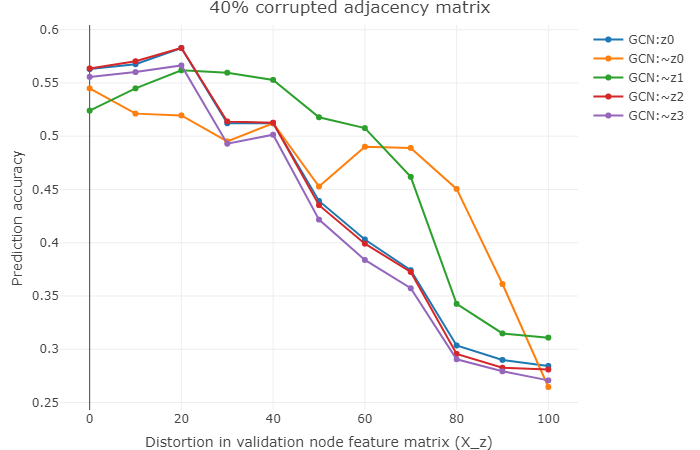}
        \subcaption{\label{fig:sub:w_GCN_X_z_A_c_40}}
    \end{minipage}
    \caption{Atypical behaviour of $\mathcal{P}(\psi_{i}^{GCN}(X_{z}^{[n_{X}]},A_{c}^{[n_{A}]}))$ and $\mathcal{P}(\phi^{GCN}(X_{z}^{[n_{X}]},A_{c}^{[n_{A}]})), i \in \{0,1,2,3\}, n_{X} \in \{0,10,\dots,100\}, n_{A} \in \{30,40\}$ for {\it WikiCS} dataset.}
    \label{fig:w_GCN_X_z_A_c}
\end{figure}

\begin{figure}[htpb]
    \centering
    \begin{minipage}{0.33\textwidth}
        \centering
        \includegraphics[scale=0.175]{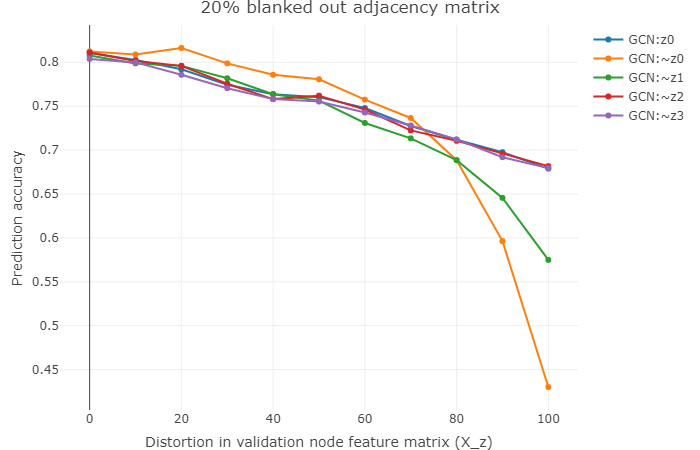}
        \subcaption{\label{fig:sub:w_GCN_X_z_A_z_20}}
    \end{minipage}%
    \begin{minipage}{0.33\textwidth}
        \centering
        \includegraphics[scale=0.175]{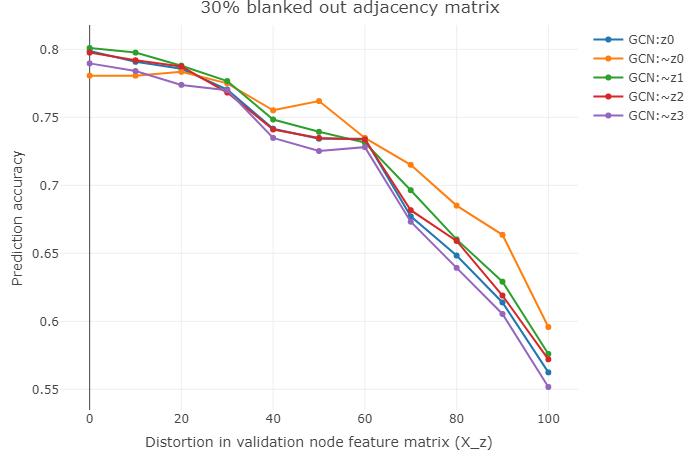}
        \subcaption{\label{fig:sub:w_GCN_X_z_A_z_30}}
    \end{minipage}%
    \begin{minipage}{0.33\textwidth}
        \centering
        \includegraphics[scale=0.175]{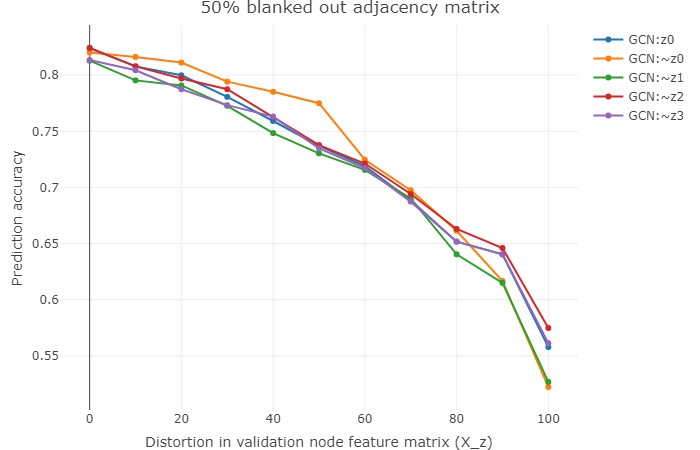}
        \subcaption{\label{fig:sub:w_GCN_X_z_A_z_50}}
    \end{minipage}
    \caption{Atypical behaviour of $\mathcal{P}(\psi_{i}^{GCN}(X_{z}^{[n_{X}]},A_{z}^{[n_{A}]}))$ and $\mathcal{P}(\phi^{GCN}(X_{z}^{[n_{X}]},A_{z}^{[n_{A}]})), i \in \{0,1,2,3\}, n_{X} \in \{0,10,\dots,100\}, n_{A} \in \{40,60,80\}$ for {\it WikiCS} dataset.}
    \label{fig:w_GCN_X_z_A_z}
\end{figure}

Figures~\ref{fig:GCN_X_c_A_c_0},~\ref{fig:w_GCN_X_c_A_c},~\ref{fig:w_GCN_X_c_A_z},~\ref{fig:o_GCN_X_z_A_c},~\ref{fig:w_GCN_X_z_A_c} and~\ref{fig:w_GCN_X_z_A_z} depict atypical behaviours of the GCN-based denoising pipelines.
~
~
~
~
~

\subsection{SAGE-based denoising pipelines}
\begin{figure}[htpb]
    \centering
    \begin{minipage}{0.33\textwidth}
        \centering
        \includegraphics[scale=0.175]{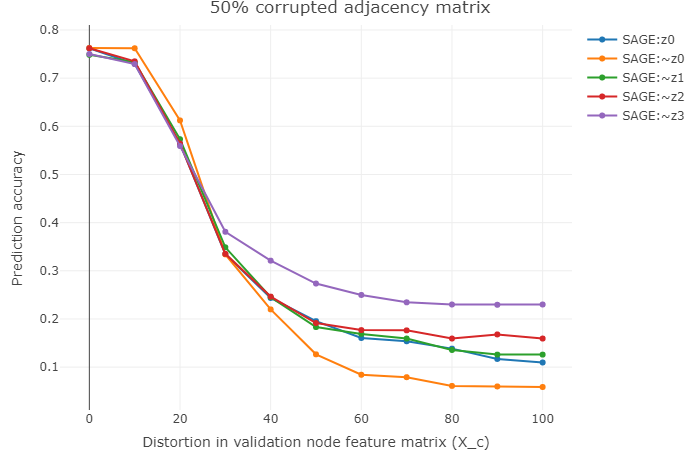}
        \subcaption{\label{fig:sub:w_SAGE_X_c_A_c_50}}
    \end{minipage}%
    \begin{minipage}{0.33\textwidth}
        \centering
        \includegraphics[scale=0.175]{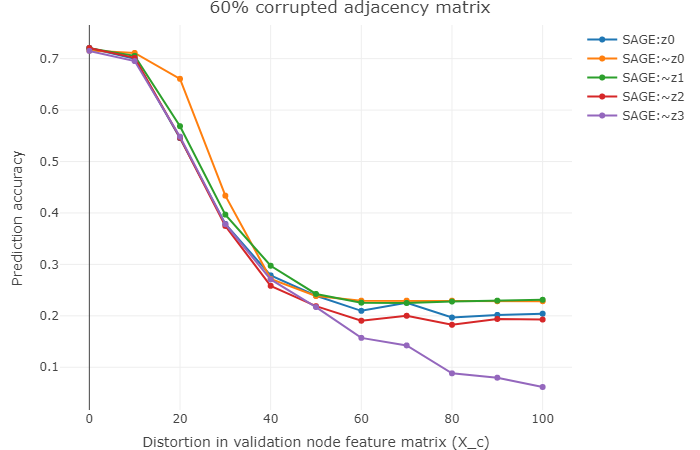}
        \subcaption{\label{fig:sub:w_SAGE_X_c_A_c_60}}
    \end{minipage}%
    \begin{minipage}{0.33\textwidth}
        \centering
        \includegraphics[scale=0.175]{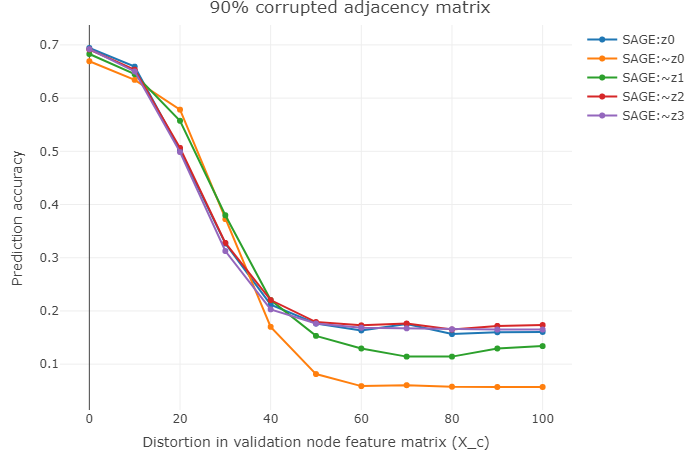}
        \subcaption{\label{fig:sub:w_SAGE_X_c_A_c_90}}
    \end{minipage}
    \caption{Atypical behaviour of $\mathcal{P}(\psi_{i}^{SAGE}(X_{c}^{[n_{X}]},A_{c}^{[n_{A}]}))$ and $\mathcal{P}(\phi^{SAGE}(X_{c}^{[n_{X}]},A_{c}^{[n_{A}]})), i \in \{0,1,2,3\}, n_{X} \in \{0,10,\dots,100\}, n_{A} \in \{50,60,90\}$ for {\it WikiCS} dataset.}
    \label{fig:w_SAGE_X_c_A_c}
\end{figure}

\begin{figure}[htpb]
    \centering
    \begin{minipage}{0.5\textwidth}
        \centering
        \includegraphics[scale=0.25]{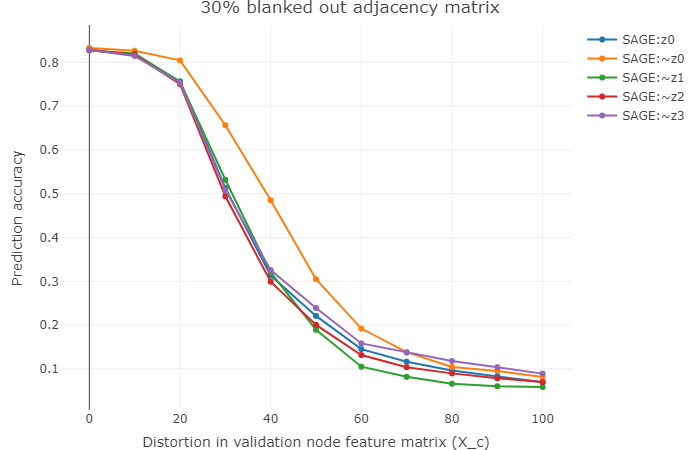}
        \subcaption{\label{fig:sub:w_SAGE_X_c_A_z_30}}
    \end{minipage}%
    \begin{minipage}{0.5\textwidth}
        \centering
        \includegraphics[scale=0.25]{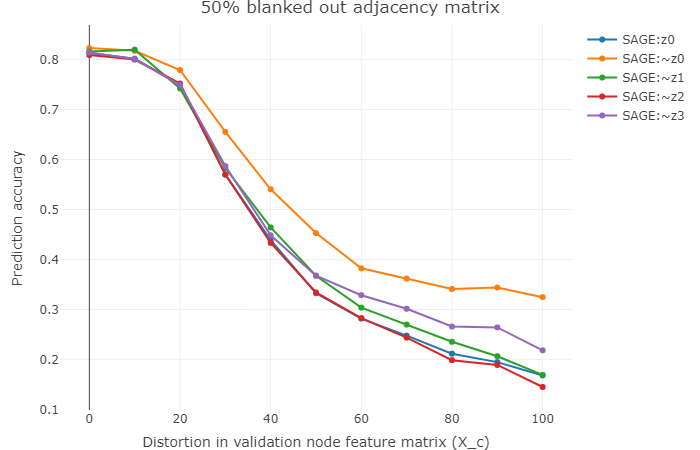}
        \subcaption{\label{fig:sub:w_SAGE_X_c_A_z_50}}
    \end{minipage}
    \caption{Atypical behaviour of $\mathcal{P}(\psi_{i}^{SAGE}(X_{c}^{[n_{X}]},A_{z}^{[n_{A}]}))$ and $\mathcal{P}(\phi^{SAGE}(X_{c}^{[n_{X}]},A_{z}^{[n_{A}]})), i \in \{0,1,2,3\}, n_{X} \in \{0,10,\dots,100\}, n_{A} \in \{30,50\}$.}
    \label{fig:SAGE_X_c_A_z}
\end{figure}

\begin{figure}
    \centering
    \includegraphics[scale=0.34]{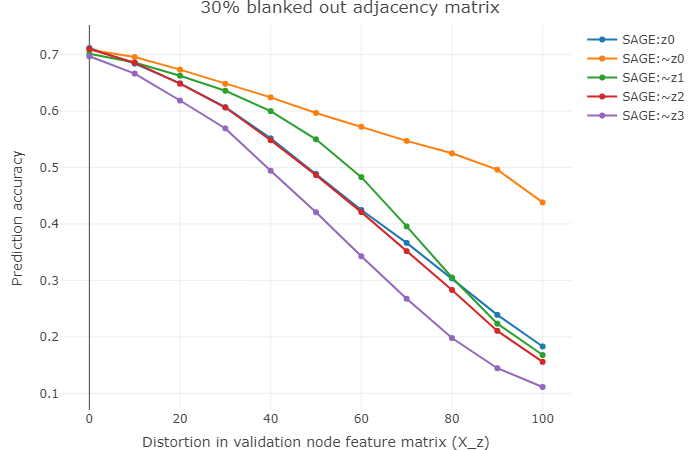}
    \caption{Atypical behaviour of $\mathcal{P}(\psi_{i}^{SAGE}(X_{z}^{[n_{X}]},A_{z}^{[30]}))$ and $\mathcal{P}(\phi^{n2v}(X_{z}^{[n_{X}]},A_{z}^{[30]})), i \in \{0,1,2,3\}, n_{X} \in \{0,10,\dots,100\}$ for {\it ogbn-arxiv} dataset.}
    \label{fig:SAGE_X_z_A_z_30}
\end{figure}

Figures~\ref{fig:w_SAGE_X_c_A_c},~\ref{fig:SAGE_X_c_A_z},~\ref{fig:SAGE_X_z_A_z_30} illustrate atypical behaviours of SAGE-based denoising pipelines.
~
~
%%%%%%%%%%%%%%%%%%%%%%%%%%%%%%%%%%%%%%%%%%%%%%%%%%%%%%%%%%%%%%
\end{document}